\documentclass{article}

\usepackage{microtype}
\usepackage{graphicx}
\usepackage{booktabs} 

\usepackage{hyperref}
\usepackage[table]{xcolor}

\usepackage{adjustbox}		
\usepackage[bottom]{footmisc}

\newcommand{\tabincell}[2]{\begin{tabular}{@{}#1@{}}#2\end{tabular}}

\newcommand{\eg}{\textit{e.g.,}~}
\newcommand{\ie}{\textit{i.e.,}~}

\newcommand{\bs}{\bm{s}}
\newcommand{\ba}{\bm{a}}
\newcommand{\bc}{{c}}
\newcommand{\btau}{\bm{\tau}}

\usepackage[accepted]{icml2023}

\usepackage{amsmath}
\usepackage{amssymb}
\usepackage{mathtools}
\usepackage{amsthm}
\usepackage{bm}
\usepackage{threeparttable}
\usepackage{multirow}
\usepackage{caption}
\usepackage{subcaption}
\usepackage{stfloats}
\usepackage[capitalize,noabbrev]{cleveref}

\theoremstyle{plain}

\theoremstyle{definition}

\theoremstyle{remark}

\definecolor{ashgrey}{rgb}{0.7, 0.75, 0.71}
\usepackage[textsize=tiny]{todonotes}

\icmltitlerunning{ChiPFormer: Transferable Chip Placement via Offline Decision Transformer}

\begin{document}

\twocolumn[
\icmltitle{ChiPFormer: Transferable Chip Placement via Offline Decision Transformer}




\begin{icmlauthorlist}
\icmlauthor{Yao Lai}{allaff}
\icmlauthor{Jinxin Liu$^{\;\textnormal{4}}$}{}
\icmlauthor{Zhentao Tang$^{\;\textnormal{3}}$}{}
\icmlauthor{Bin Wang$^{\;\textnormal{3}}$}{}
\icmlauthor{Jianye Hao$^{\;\textnormal{3\ 5}}$}{}
\icmlauthor{Ping Luo$^{\;\textnormal{2\ 1}}$}{}
\end{icmlauthorlist}

\icmlaffiliation{allaff}{
Department of Computer Science, The University of Hong Kong, Hong Kong 
$^\textnormal{2}$ Shanghai AI Laboratory, China 
$^\textnormal{3}$ Huawei Noah’s Ark Lab, China 
$^\textnormal{4}$ Zhejiang University, China
$^\textnormal{5}$ Tianjin University, China
}

\icmlcorrespondingauthor{Ping Luo}{pluo@cs.hku.hk}
\icmlcorrespondingauthor{Jianye Hao}{jianye.hao@tju.edu.cn}

\icmlkeywords{Machine Learning, ICML}

\vskip 0.3in
]



\printAffiliationsAndNotice  

\begin{abstract}
Placement is a critical step in modern chip design, aiming to determine the positions of circuit modules on the chip canvas.
Recent works have shown that reinforcement learning (RL)  can improve human performance in chip placement. 
However, such an RL-based approach suffers from long training time and low transfer ability in unseen chip circuits. 
To resolve these challenges, we cast the chip placement as an offline RL formulation and present ChiPFormer that enables learning a transferable placement policy from fixed offline data. 
ChiPFormer has several advantages that prior arts do not have.
First, ChiPFormer can exploit offline placement designs to learn transferable policies
more efficiently in a multi-task setting. 
Second, ChiPFormer can promote effective finetuning for unseen chip circuits, reducing the placement runtime from hours to minutes. 
Third, extensive experiments on 32 chip circuits demonstrate that ChiPFormer achieves significantly better placement quality while reducing the runtime by 10$\times$ compared to recent state-of-the-art approaches in both public benchmarks and realistic industrial tasks.
The deliverables are released at \href{https://sites.google.com/view/chipformer/home}{sites.google.com/view/chipformer/home}.
\end{abstract}

\section{Introduction}

\begin{figure}[t]
\begin{center}
\centerline{\includegraphics[width=0.8\columnwidth]{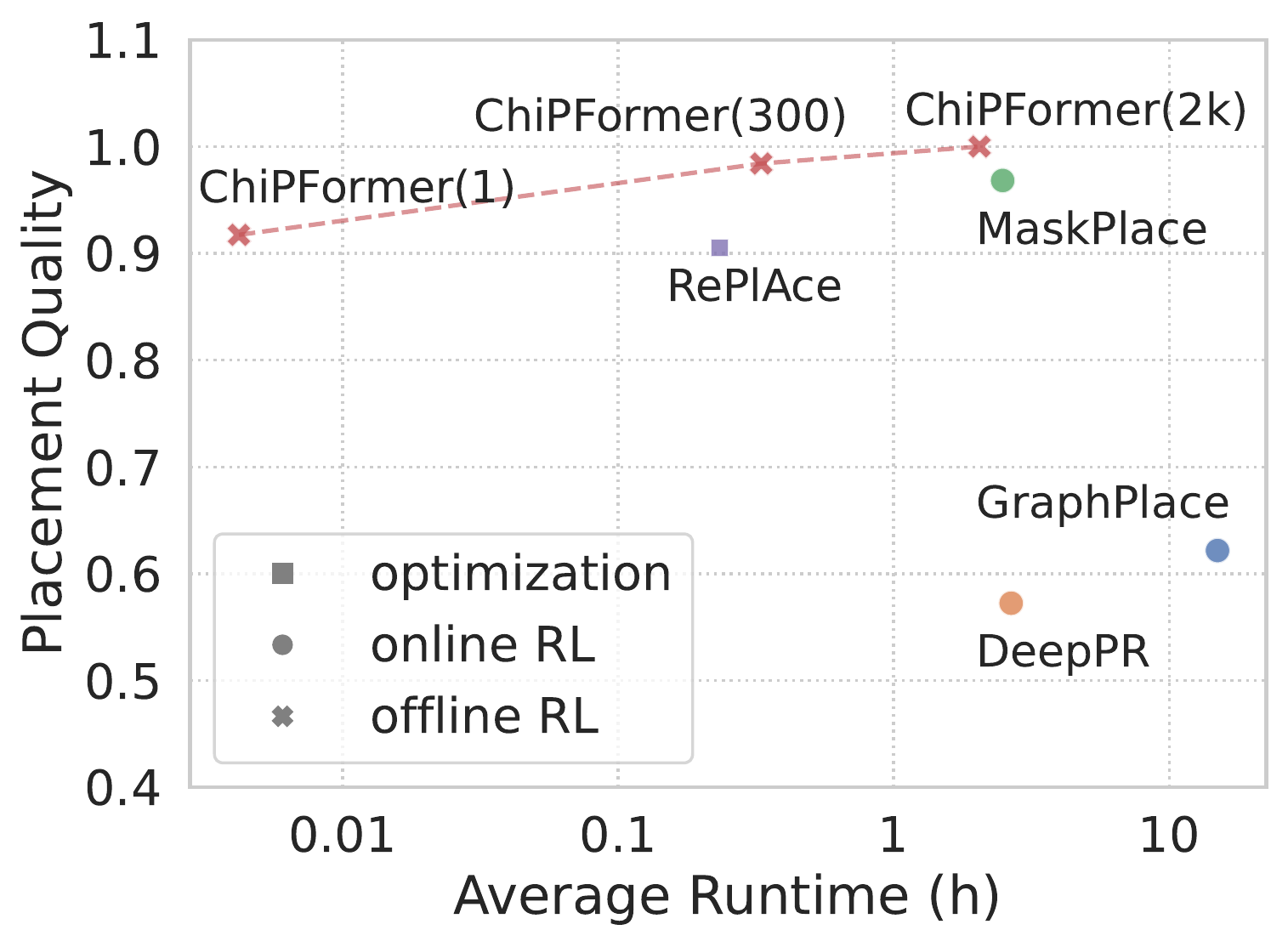}}
\caption{\textbf{Comparing the placement quality and the runtime} between  ChiPFormer (ours) and the recent advanced RL-based methods such as GraphPlace \cite{mirhoseini2021graph}, MaskPlace \cite{lai2022maskplace}, and DeepPR \cite{cheng2021joint} and optimization method RePlAce \cite{cheng2018replace}. All methods are evaluated in the \textit{ISPD05} benchmark.
The placement quality (higher is better) is normalized by $1.25^{(1-\text{HPWL}/\min\ \text{HPWL})} \in [0, 1]$, where \text{HPWL} represents Half Perimeter Wire Length. The number 
inside ``ChiPFormer($\cdot$)'' means the maximum few-shot number  for placement in the finetuning stage (\ie rollout times). We see that ChiPFormer is the first offline RL approach for chip placement so far, and ChiPFormer($300$) outperforms the other baselines in terms of quality and efficiency.
}
\label{comparison}
\end{center}
\end{figure}

\begin{figure*}[!ht]
\begin{center}
\centerline{\includegraphics[width=2\columnwidth]{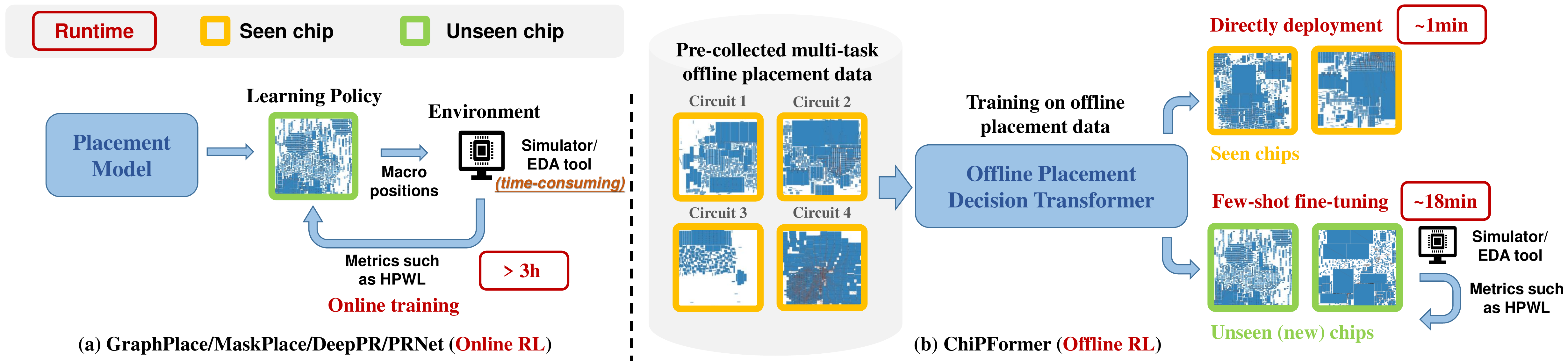}}
\caption{\textbf{Comparing the overall pipelines between  (a) online RL placement and (b) offline RL placement (ours).}
In (a), the online RL model  keeps interacting with the environment (\ie a placement simulator or an EDA design tool for getting metrics from placement designs, and its time consumption is usually proportional to the circuit scale) to learn policy from scratch. As a result, online policy learning takes more than 3 hours on a chip in the \textit{ISPD05} benchmark. 
In (b), the offline approach enables ChiPFormer to learn a policy using fixed offline placement data, thus eliminating time-consuming online interactions. 
When an \textit{unseen} circuit is presented, ChiPFormer can be finetuned with only a few rollouts by reusing and transferring the learned experience across multi-task offline data. We see that such an offline method can reduce the runtime in the \textit{ISPD05} benchmark ten times compared to previous online approaches (\ie 18 minutes \textit{v.s.} 3 hours).
}
\label{process}
\end{center}
\end{figure*}

In modern chip design, placement is a crucial and challenging problem, which places circuit modules (\eg macros and standard cells) with varying sizes on a  chip canvas. 
The placement result can determine a chip's performance, such as the speed and energy cost, especially when the scale of integrated circuits grows continuously. 
For example, very large-scale integrated (VLSI) circuits can have $100$ to $1$k  macros (\eg SRAMs, IOs, and packaged computing units) and $10$k to $100$k standard cells (\eg logical gates), making the placement problem computationally expensive.

Recent advanced approaches \cite{mirhoseini2021graph, cheng2021joint, lai2022maskplace} have shown that reinforcement learning (RL)  can produce chip layouts that are superior or comparable to those designed by humans in many key evaluation metrics such as wirelength and congestion while 
spending less placement time than humans.
Specifically, these approaches typically treat the placement problem as a Markov Decision Process (MDP) and place one circuit module at each step.
All of them learn the placement policy in an online manner by iteratively collecting data through interactions with the environment, as shown in Fig.\ref{process}~(a).

However, one challenge that remains unsolved is that all recent RL methods have a low training efficiency, as shown in Fig.\ref{comparison}. This is because learning the policy online is slow in a large placement search space, significantly when the chip circuit scale increases. 
For instance, \citet{mirhoseini2021graph} pointed out that the search space can be larger than $10^{2500}$ when there are only $1000$ modules. In that case, it takes approximately $48$ hours to pretrain a policy network and $6$ hours to finetune it on eight  V$100$ GPUs. Although DeepPR~\citep{cheng2021joint} and MaskPlace~\citep{lai2022maskplace} can shorten the runtime, they still require  more than $3$ hours of training. 
RL-based methods can achieve much stronger placement performance than the classic optimization approaches \cite{lin2020dreamplace,cheng2018replace}, but their long runtime makes them less practical than classic methods that can produce a placement design in minutes.

This work addresses the above challenge by presenting ChiPFormer, which designs a decision transformer model for chip placement. 
It has three appealing benefits compared to the recent approaches. 
First, unlike the current advanced RL methods that learn the placement policy online, ChiPFormer formulates chip placement  as an offline RL problem in a data-driven manner. 
This offline formulation enables us to pretrain ChiPFormer on fixed and pre-collected data, alleviating the time-consuming online rollouts and enabling data reuse across 
multiple placement tasks. The learned policy can be transferred to unseen chip circuits in a few~minutes.

Second, ChiPFormer can learn transferable placement policy, which is achieved by collecting {multi-task} offline data on multiple circuits and then modeling a conditional placement policy with ChiPFormer.
Following Gato~\citep{reed2022generalist}, we assume the offline data are collected from multiple chip circuits using (near) expert-level placement behaviors, differing from the common offline RL setting where the data are collected by sub-optimal behavior policies over a {single} environment such as~\citet{chen2021decision}. 

Third, unlike the recent representative RL methods such as GraphPlace \cite{mirhoseini2021graph}, MaskPlace \cite{lai2022maskplace}, and DeepPR \cite{cheng2021joint} that learned the placement policy using convolutional neural network (CNN) or graph neural network (GCN), ChiPFormer is the first transformer placement network so far. Our multi-task transformer design allows us to learn transferable policies, which can be generalized to new unseen circuits
within a few minutes. 
For example, we apply ChiPFormer on $12$ unseen chip circuits and achieve an average placement time of 18 minutes, outperforming the GraphPlace method \cite{mirhoseini2021graph} in HPWL and runtime metrics by $65$\% and $97$\% decrease, respectively.

This paper has {three} main \textbf{contributions}.
(1) To our knowledge, ChiPFormer is the first work so far to learn transferable placement policy in an offline RL manner. The learned policy can generalize to unseen chip placement tasks effectively and efficiently. 
(2) To facilitate further study of the offline placement problem, we  release the {collected} placement dataset, including $12$ chip circuits (tasks) and $500$ {expert} placement results for each circuit.\footnote{The dataset is  shared on \href{https://drive.google.com/drive/folders/1F7075SvjccYk97i2UWhahN_9krBvDCmr?usp=sharing}{Google drive}.}
(3) We conduct extensive experiments on $32$ circuit tasks, containing $26$ circuits from public chip benchmarks and $6$ circuits from {realistic} industrial chips. This work has evaluated $1.3\times \sim 5.3\times$ more chip circuits than recent works.
For example, there are $6$ circuits evaluated in GraphPlace, $24$ in MaskPlace, and $8$ in DeepPR.
In all experiments, ChiPFormer can speed up the placement runtime by $10\times\sim30\times$~with only two Nvidia 3090 GPUs while achieving better placement quality compared to the recent state-of-the-art methods. 

\section{Preliminaries}

\textbf{Chip Placement.} 
The chip placement problem can be defined as a constrained optimization problem. The main objective is to determine the positions of circuit modules (\eg macros and standard cells) on a physical chip canvas
to minimize the wirelength, which determines the chip delay and energy consumption. 
As the computation of wirelength is an NP-complete problem~\citep{garey1977rectilinear}, recent placement methods use Half-Perimeter Wire-Length (HPWL) as a proxy to estimate the wirelength, which is computed by accumulating all the half-perimeters of bounding boxes of all the nets from the circuit netlist. 

As shown below, the  constraints for placement include: (1) \textit{overlap constraint}, which avoids the overlapping between modules (\ie each position on the chip canvas can be covered by at most one module), and (2) \textit{congestion constraint}, where the wire congestion should be lower than a desired small threshold to reduce chip cost. 
In general, the placement optimization problem can be formulated as
\begin{align}
\min_{\bm{x}, \bm{y}}&\  \text{HPWL}(\bm{x}, \bm{y}),\\ 
\text{s.t.}&\ \text{Overlap}(\bm{x}, \bm{y}, \bm{w}, \bm{h})=0,\\ 
\quad\ &\ \text{Congestion}(\bm{x}, \bm{y}, \bm{w}, \bm{h})\leq C,
\label{e1}
\end{align}
where $(\bm{x}, \bm{y}) = (x_1, x_2, ..., x_n, y_1, y_2, ..., y_n)$ and each pair $(x_i, y_i)$ is the placement position of the $i^{th}$ circuit module.  Similarly, $(\bm{w}, \bm{h})$ are the widths and heights of modules. $C$ is the desired threshold of the congestion constraint. $\text{HPWL}(\cdot)$, $\text{Overlap}(\cdot)$, and $\text{Congestion}(\cdot)$ are the functions to calculate the HPWL, the overlap area, and the congestion of the placement design, respectively. 

\textbf{Offline Reinforcement Learning.} 
RL is typically designed to deal with tasks of sequential modeling, which is often described by the Markov Decision Process (MDP). Particularly, an MDP, written as $\mathcal{M}:=(\mathcal{S}, \mathcal{A}, \mathcal{T}, {r}, \rho, \gamma)$, is specified by the state space $\mathcal{S}$, action space $\mathcal{A}$, transition dynamics $\mathcal{T}(\bs_{t+1}|\bs_t,\ba_t)$, reward function $r(\bs_t,\ba_t)$, initial
state distribution $\rho(\bs_1)$, and discount factor $\gamma$. The goal of RL is to learn a policy $\pi(\ba|\bs)$ that maximizes the expected return $\mathbb{E}_{\btau\sim \pi(\btau)}[\sum_{t=1}^Tr(\bs_t, \ba_t)]$, where $\btau:=(\bm{s}_1, \bm{a}_1, ..., \bm{s}_T, \bm{a}_T)$ denotes the state-action trajectory. 
For clarity, we employ the notation $\pi(\btau)$ to denote the trajectory distribution induced by executing the policy $\pi(\ba|\bs)$ in the environment. 

In \textit{offline} RL, the agent (learning policy) will not interact with the environment (\ie it cannot input the placement design into the simulator/EDA tool and get the metrics). Instead, the agent is provided with a fixed dataset $\mathcal{D}:=\{(\bs, \ba, \bs', r)\}$, which has been collected by some data-generating process. Since it cannot explore the MDP, the agent must rely on the provided offline data to learn a policy that maximizes the expected return.

\textbf{Reinforcement Learning for Placement.} To instantiate an RL-based chip placement, we can re-frame the placement as a sequential decision-making MDP, denoted by $\mathcal{M}^{c}$, where we use the superscript to indicate the chip placement task for a specific circuit $c$ \cite{mirhoseini2021graph, lai2022maskplace}. 
In the MDP, state $\bs$ describes the positions of previous macros that have been placed, action $\ba$ indicates the position of the current macro to be placed, and reward $r$ is defined as the negative wirelength and constraints (circuit-dependent).
\section{Our Approach}

\subsection{Problem Setup}
This paper studies the chip placement problem in the offline RL regime, as shown in Fig.\ref{process} (b). 
In particular, we cast the chip placement as a sequential decision-making problem but assume that the RL experience is fixed and there is no further interaction with the environment, which has a vast search space.  
In contrast, previous RL placement methods learn from scratch through expensive and time-consuming online interaction with the environment, creating a barrier to applying RL methods to {efficient} placement tasks.
Hence, we expect that offline training would facilitate more efficient chip placement, especially when transferring to new unseen chip circuit tasks.

Formally,  we have an \textit{expert-level} and \textit{multi-task} chip placement dataset, denoted as  $\mathcal{D}:=\{(\bc, \btau)\}$, where $\bc$ denotes the index of a chip circuit task and $\btau$ denotes the collected expert-level placement behaviors corresponding to $\bc$. 
Unlike the vanilla offline RL, we drop the reward term by considering expert-level behaviors. 

Different from typical offline RL that assumes  all offline data comes from a single task, our offline data $\mathcal{D}$ are collected from $n$ different circuit placement tasks $\mathcal{M}_{\text{train}}^{[n]}$, where $\mathcal{M}_{\text{train}}^{[n]}$ denotes a set of MDPs $\{\mathcal{M}_{\text{train}}^1, \cdots, \mathcal{M}_{\text{train}}^{n}\}$. This is conceptually similar to the multi-task learning setup. 
In modern chip placement, the {expert-level} and {multi-task} setup  can be naturally satisfied. For example, one can employ any RL-based and
optimization-based methods to collect the  placement data from multiple existing circuits. To facilitate future research, we have released our collected offline {dataset}.

\begin{figure*}[!htb]
\begin{center}
\centerline{\includegraphics[width=2\columnwidth]{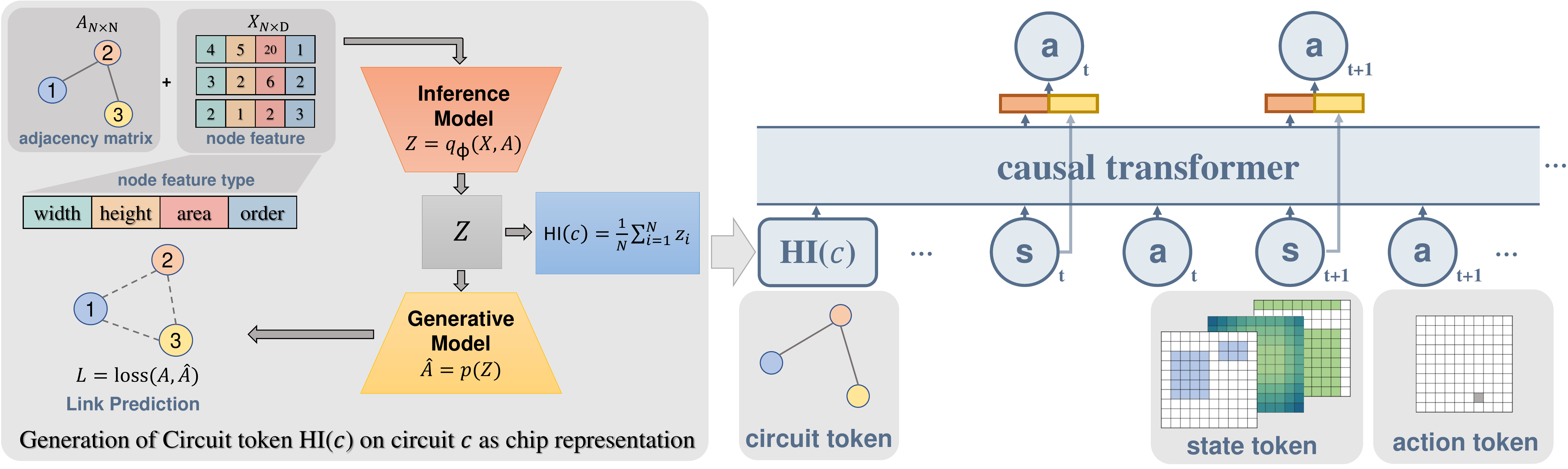}}
\caption{\textbf{Circuit token generation (left) and ChiPFormer architecture (right).} 
We generate the circuit token $\text{HI}(c)$ by pretraining a conditional VAE model (graph implementation), which fits an inference model and a generative model. Specifically, the inference model takes as input the adjacency matrix $\textbf{A}$ and node feature matrix $\textbf{X}$ of a graph representing the topology structure of circuit~$\bc$. 
We use the GPT architecture as the backbone of ChiPFormer, which uses a causal self-attention mask and predicts actions by feeding tokens (including circuit token, state token, and action token) autoregressively.}
\label{arch}
\end{center}
\end{figure*}

\subsection{Overall Pipeline}
Our pipeline consists of two steps, including  macro placement and mixed-size 
placement. 

In the first step for macro placement,
we train a general decision/sequence generation model, ChiPFormer, using only the collected data $\mathcal{D}$ of multiple placement tasks. Then, we transfer the learned policy to a downstream task (chip), denoted by $\mathcal{M}_{\text{test}}^{ }$. 
For example, as Fig.\ref{process} (b) shows, 
given an already seen chip, \ie the downstream task stays in the offline training distribution $\mathcal{M}_{\text{test}}\in 
\mathcal{M}_{\text{train}}^{[n]}$,
we can directly use the pretrained ChiPFormer  to generate a new placement design. 
Alternatively, given a new unseen chip, \ie $\mathcal{M}_{\text{test}}\notin 
\mathcal{M}_{\text{train}}^{[n]}$,
we can finetune the pretrained ChiPFormer  for adapting to unseen circuits with few-shot online interaction (Section~\ref{sec:online-finetune}). 
In the second step for mixed-size placement, we let the macro placement result as an initial layout status and adopt the optimized-based method to place the standard cells (Section~\ref{sec:abm-standard-cell-placement}) 
by following recent works such as Flora \cite{liu2022floorplanning} and GraphPlanner \cite{liu2022graphplanner}.

\subsection{Offline Learning of ChiPFormer}
\label{sec:offline-chipformer}

ChiPFormer models the sequential placement task as a general {hindsight information matching}~\citep{furuta2021generalized}, by following prior supervised offline RL methods~\citep{chen2021decision, emmons2021rvs}.
We first describe its objective and then present the details of the network architecture for sequential placement modeling.

\textbf{Hindsight Information Matching.} 
Specifically, given a trajectory $\bm{\tau}:=(\bs_1, \ba_1, \cdots, \bs_{T}, \ba_{T})$ starting from state $\bs_1$, we can define \textit{hindsight information} as a trajectory's information statistics, denoted as $\text{HI}(\btau)$, which could be any function of a trajectory that captures some statistical properties in the state space or the trajectory space. For example, a prior work, Decision Transformer~\citep{chen2021decision}, takes the return of a trajectory as the {hindsight information} $\text{HI}(\bm{\tau}) := \sum_{t=1}^{T}r(\bm{s}_t, \bm{a}_t)$. The supervised goal-reaching RL~\citep{ghosh2019learning} takes the final state as the hindsight information $\text{HI}(\btau) := \bs_{T}$. 
In contrast, we extend the naive {hindsight information} to the \textit{multi-task} scenario by incorporating the additional task description  (\ie circuit $c$). Instead of using heuristic statistics (like the returns), we take a learned embedding of the circuit $\bc$ as the multi-task {hindsight information} $\text{HI}(\bc, \btau)$. 

We then model the above hindsight information matching  via supervised regression by learning a conditional placement policy $\pi_\theta$, 
\begin{align}
\max_{\theta} \ \mathbb{E}_{(c,\btau)\sim \mathcal{D}}\left[\sum_{t=1}^{T} \log \pi_\theta(\ba_{t}|\btau_{t}, \text{HI}(c,\btau)) \right],\label{eq:hindsight-information-matching}
\end{align}
where $\btau_{t}:= (\bs_1, \ba_1, \cdots, \bs_t)$ denotes the trajectory segment  until the state $\bs_t$. 
Next, we discuss how such placement {hindsight information} $\text{HI}(c, \btau)$ can be captured by exploiting the circuit netlist structure as its representation.

\textbf{Chip Representation.} Since the physical connectivity of the circuit netlist can often be represented by a graph, we use the variational graph auto-encoders (VGAE)~\citep{kipf2016variational} 
to encode the topology information of chip circuits.  
As shown in the left of Fig.\ref{arch}, 
we transform the netlist of the circuit $\bc$ into an undirected graph $\mathcal{G}^\bc = (\mathcal{V}^\bc,\mathcal{E}^\bc)$, where $\mathcal{V}^\bc$ denotes a set of nodes representing all modules on a circuit $\bc$, and $\mathcal{E}^\bc$ denotes a set of edges representing the wire connecting the modules. 
We also define an $N\times N$ matrix $\textbf{A}^\bc$ and an $N\times D$ matrix $\textbf{X}^\bc$ to be the adjacency matrix and the node features of graph $\mathcal{G}^\bc$, respectively.
We have $N = |\mathcal{V}^c|$, the number of nodes, and $D$, the size of the node features. 

Then, we can optimize the following variational lower bound with respect to the inference parameters $\phi$ as shown below. For simplicity of notation, here we drop the superscript $\bc$ for matrices $\textbf{X}^\bc$, $\textbf{A}^\bc$, $\textbf{Z}^\bc$, and vector $\bm{z}^\bc$.
\begin{align}
\max_{\phi} \ \mathbb{E}_{\bc\sim\mathcal{D}} &\left[\mathbb{E}_{q_\phi(\bm{Z}|\bm{X}, \bm{A})}[\text{log}\ p(\bm{A}|\bm{Z})] \right. \nonumber\\
&\left. - \text{KL}[q(\bm{Z}|\bm{X},\bm{A})||p(\bm{Z})] \right], \label{eq:vgae-objective}
\end{align}
where inference model $q_\phi(\bm{Z}|\bm{X}, \bm{A}) = \prod_{i=1}^{N} q_\phi(\bm{z}_i|\bm{X},\bm{A})$, and generative model $p(\bm{A}|\bm{Z}) = \prod_{i=1}^N\prod_{j=1}^N\sigma(\bm{z}_i^\text{T}\bm{z}_j)$. In addition, we have $\bm{Z} = [\bm{z}_1, \bm{z}_2, \cdots, \bm{z}_N]^\text{T}$
as the latent variables to represent nodes. $p(\bm{Z})$ is a Gaussian prior, and $\sigma(x) = 1/(1+e^{-x})$ is the sigmoid function. $\text{KL}[p(\cdot)||q(\cdot)]$ is the Kullback-Leibler divergence  between $p(\cdot)$ and $q(\cdot)$.

As we have collected an expert-level offline dataset, we can remove the dependency of the hindsight information $\text{HI}(\bc,\btau)$ on the expert trajectory $\btau$, thus replacing $\text{HI}(\bc,\btau)$ with $\text{HI}(\bc)$ and approximating $\text{HI}(\bc)$ through the learned surrogate inference model ${q_\phi}$ in Eqn.\eqref{eq:vgae-objective}. 

Following prior work \cite{mirhoseini2021graph}, we compute the mean of all latent embeddings 
to obtain a representation for the whole graph $\mathcal{G}^\bc$, \ie setting the offline placement hindsight information $\text{HI}(c,\btau) = \text{HI}(c) = \frac{1}{N}\sum_{i=1}^{N}\bm{z}_i^\bc$, 
where $\bm{z}_i^\bc$ is a row vector of the matrix $\bm{Z}^\bc = q_\phi(\bm{Z}^\bc|\bm{X}^\bc, \bm{A}^\bc)$. The corresponding pseudo-code is shown in Appendix Algo. \ref{alg-circuit-token}.

\textbf{Network Architecture.} 
We illustrate the architecture of  ChiPFormer  at the right of Fig.\ref{arch}, which uses GPT~\citep{radford2018improving, radford2019language} as the backbone network. 
Specifically, there are $(2T+1)$ tokens in one placement trajectory, including $T$ state tokens, $T$ action tokens, and one circuit token to represent the hindsight information $\text{HI}(\bc)$.

We determine the placement order
by the sorting results of the area and the net number of the macros. 
We represent the state token $\bs_t$ using a three-channel input, including the position mask, the wire mask, and the view mask, similar to MaskPlace~\citep{lai2022maskplace}. These mask representations are discussed in Appendix \ref{state-token}. To scale the input size, we divide the chip canvas into an $84\times 84$ grid. Thus each input channel can be seen as an image with the resolution $84\times 84$, and we can represent the state as a three-channel input of size $3\times84\times 84$. 
For the action token $\ba_t$, we represent it as the two-dimensional $(x,y)$ coordinates of the macro $t$, which will be placed in the 2D grid. To avoid overlapping between macros, we also remove all infeasible actions by the position mask introduced in Appendix \ref{state-token}.

Before feeding the state and the action tokens into the GPT backbone, we employ 
two trainable embedding models to first encode the states and actions separately. 
Furthermore, we introduce a circuit token to distinguish different circuit topologies. The main reason is that the visual representation in the state token ignores topology information of circuits, which is important for  placement, especially in multi-task learning. If two circuits contain the same modules but are connected differently, the optimal placement solution should be different as Appendix \ref{didactic-example}. Thus, we encode the circuit topology information into the hindsight information (Eqn.\eqref{eq:hindsight-information-matching}). 
As the circuit token, $\text{HI}(\bc)$ is independent of the placement state token and should be considered in all sequence modeling steps. Therefore, it is put at the beginning of the sequential modeling process.  Detailed model architecture and hyper-parameter settings are in Appendix \ref{architecture-and-hyper}, Table \ref{model-arch-tab} and \ref{configuration-tab}.

\subsection{Online Finetuning for Unseen Chip}
\label{sec:online-finetune}

For a new unseen chip placement task (\ie $\mathcal{M}_{\text{test}}\notin \mathcal{M}_{\text{train}}^{[n]}$), we can finetune the pretrained ChiPFormer model by few-shot online rollouts similar to the online decision transformer \cite{zheng2022online}. To maintain the training stability of the supervised pretraining process, we finetune the policy via a return-weighted regression through online interaction with the unseen {circuit $c_{\text{test}}$}: 
\begin{align}
\min_{\theta} \ \mathcal{L}(\pi_\theta) := -  \mathbb{E}_{\mathcal{B}(\btau)} \left[ \omega(\btau) \log \pi_\theta(\ba_t|\btau_{t}, \text{HI}(\bc_{\text{test}}))\right], \nonumber 
\end{align}
where the expectation is over the replay buffer $\mathcal{B}$ of online rollout trajectories, $\omega(\btau):= \frac{\exp(R(\btau)/\alpha)}{\mathbb{E}_{\mathcal{B}(\btau)}\left[\exp(R(\btau)/\alpha)\right]}$ is the return-guided prioritization weight, $R(\btau) := \sum_{t=1}^{T}r(\bs_t,\ba_t)$ is the return of a trajectory, and $\alpha$ is the sampling temperature. 
In the implementation, we also build the online replay buffer $\mathcal{B}$ on a priority queue to keep the trajectories with the highest returns in the search history instead of the FIFO (First-In First-Out) replay buffer in \citet{zheng2022online}.

Further, we employ a policy entropy as an intrinsic motivation $\mathcal{H}(\pi_\theta) = \mathbb{E}_{\mathcal{B}(\btau)} \left[ - \log \pi_\theta(\cdot|\bs_t, \text{HI}(\bc_{\text{test}}))  \right]$,  encouraging the online rollout to yield more exploratory behaviors. Combining the weighted regression and the entropy objective, we obtain the following  policy finetuning objective:
\begin{align}
\min_{\theta} \ \mathcal{L}(\pi_\theta)  + \lambda \max(0, \beta - \mathcal{H}(\pi_\theta)), \label{eq:finetuning}
\end{align}
where $\lambda$ is the hyper-parameter to represent the weight of the entropy loss, and we introduce the $\max$ operator to keep the value of entropy $\mathcal{H}(\pi_\theta)$ larger than one threshold $\beta$, so as to maintain a certain level of exploration.

\begin{table}[!htbp]
	\caption{{\textbf{Mixed-size placement workflow.} In the implementation, we adopt DREAMPlace~\citep{lin2020dreamplace} to conduct the optimization-based placement procedure.}}
	\label{placementflow}
	\vspace{-11pt}
	\begin{center}
		\begin{small}
			\begin{adjustbox}{max width=1.0\columnwidth}
			\begin{threeparttable}
				\begin{tabular}{ccl}
					\toprule
					Phase & Operation\tnote{1} & Description \\
					\midrule
					1 & - & \tabincell{l}{Return the macro placement result from\\the pretrained (or finetuned) ChiPFormer.}\\
					\midrule
					2  & GP & \tabincell{l}{Fix all macros placed by ChipFormer, and \\ return a coarse layout 
                    by optimization method.
                    }\\
					\midrule
					3  & GP + LG + DP & \tabincell{l}{Set macros and standard cells movable, and \\return a global optimal placement layout.} \\
					\bottomrule
				\end{tabular}
				\begin{tablenotes}
					\scriptsize
					\item[1] Operation in optimization-based placement method (DREAMPlace, \citet{lin2020dreamplace}): \\ GP = Global Placement, LG = Legalization, DP = Detailed Placement.
				\end{tablenotes}
		\end{threeparttable}
		\end{adjustbox}
		\end{small}
	\end{center}
\vspace{-8pt}
\end{table}

\subsection{Mixed-size Placement}
\label{sec:abm-standard-cell-placement}
The mixed-size placement aims at placing all modules on the chip canvas, including macros and standard cells. Our work follows similar procedures in \citet{mirhoseini2021graph, cheng2021joint}, which employ the optimized-based method for standard cell placement. 
Different from these methods that fix the positions of macros and subsequently search remaining positions for standard cells, we let the positions of macros (after macro placement) as the initial layout and allow macros still to be {movable} when performing optimization-based placement as Table~\ref{placementflow}, thus reaching the  optimal solution.

\begin{table*}[!htb]
\caption{
\textbf{Comparisons of HPWL ($\times$$10^5$) for macro placement (lower is better).}
All results are based on the performance of unseen circuits.
The number after the method name corresponds to the online rollout {times}, \ie the number of placement attempts. In particular, ChiPFormer(1) denotes the zero-shot placement performance. The percentage values in \textcolor{magenta}{pink} or \textcolor{cyan}{cyan} indicate the \textcolor{magenta}{reduction} or \textcolor{cyan}{increase} rates of HPWL compared to state-of-the-art baseline results (that have been \underline{underlined}). 
By reducing the rollout times by 10 times compared to the best baseline MaskPlace(3k), our ChiPFormer(300) can still produce the minimal wirelength in 10 out of 12 circuits. 
Further, ChiPFormer(2k) can achieve state-of-the-art results in all circuits when increasing the rollout times to 2k (which is still smaller than the one required by the previous baselines).
}
\label{macro-placement}
\vskip -0.2in
\begin{center}
\begin{small}
\begin{adjustbox}{max width=2.05\columnwidth}
\begin{tabular}{cccc|ccc}
\toprule circuit & GraphPlace(50k) & DeepPR(3k) &  MaskPlace(3k) & ChiPFormer(1) & ChiPFormer(300)  & ChiPFormer(2k) \\
\midrule
adaptec1   & 30.01$\pm$2.98 & 19.91$\pm$2.13 & \underline{7.62$\pm$0.67} & 8.87$\pm$0.98 & 
 7.02$\pm$0.11 \textcolor{magenta}{\scriptsize{(-7.87\%)}} & \textbf{6.62$\pm$0.05} \textcolor{magenta}{\scriptsize{(-13.12\%)}}\\
adaptec2   & 351.71$\pm$38.20 & 203.51$\pm$6.27 &  \underline{75.16$\pm$4.97} & 122.37$\pm$22.61 &
70.42$\pm$2.67 \textcolor{magenta}{\scriptsize{(-6.30\%)}} & \textbf{67.10$\pm$5.46} \textcolor{magenta}{\scriptsize{(-10.72\%)}}\\
adaptec3   & 358.18$\pm$13.95 & 347.16$\pm$4.32 &  \underline{100.24$\pm$13.54} & 107.11$\pm$8.84 & 
78.32$\pm$2.03 \textcolor{magenta}{\scriptsize{(-21.87\%)}} & \textbf{76.70$\pm$1.15} \textcolor{magenta}{\scriptsize{(-23.48\%)}}\\
adaptec4   & 151.42$\pm$9.72 & 311.86$\pm$56.74 &  \underline{87.99$\pm$3.25} & 85.63$\pm$7.52 & 
69.42$\pm$0.54 \textcolor{magenta}{\scriptsize{(-21.10\%)}} & \textbf{68.80$\pm$1.59} \textcolor{magenta}{\scriptsize{(-21.81\%)}}\\
bigblue1   & 10.58$\pm$1.29 & 23.33$\pm$3.65 & \underline{3.04$\pm$0.06} & 3.11$\pm$0.03 &
2.96$\pm$0.04 \textcolor{magenta}{\scriptsize{(-2.63\%)}} & \textbf{2.95$\pm$0.04} \textcolor{magenta}{\scriptsize{(-2.96\%)}}\\
bigblue2   & 14.78$\pm$0.95 & 11.38$\pm$0.20 & \underline{5.75$\pm$0.11} & 6.85$\pm$0.26  &
6.02$\pm$0.09  \textcolor{cyan}{\scriptsize{(+4.70\%)}}& \textbf{5.44$\pm$0.10} \textcolor{magenta}{\scriptsize{(-5.39\%)}}\\
bigblue3   & 357.48$\pm$47.83 & 430.48$\pm$12.18 & \underline{90.04$\pm$4.83} & 131.78$\pm$17.36 &
81.48$\pm$4.83 \textcolor{magenta}{\scriptsize{(-9.51\%)}} & \textbf{72.92$\pm$2.56} \textcolor{magenta}{\scriptsize{(-19.01\%)}}\\
bigblue4   & 440.70$\pm$15.95 & 433.90$\pm$5.26 & \underline{103.26$\pm$2.69} & 136.79$\pm$13.93 &
110.10$\pm$0.23 \textcolor{cyan}{\scriptsize{(+6.62\%)}} & \textbf{102.84$\pm$0.15} \textcolor{magenta}{\scriptsize{(-0.41\%)}}\\
ibm01   & 4.12$\pm$0.10 & 6.62$\pm$0.27 & \underline{3.73$\pm$0.12} & 4.57$\pm$0.27 &
3.61$\pm$0.08 \textcolor{magenta}{\scriptsize{(-3.22\%)}} & \textbf{3.05$\pm$0.11} \textcolor{magenta}{\scriptsize{(-18.23\%)}}\\
ibm02   & 4.56$\pm$0.03 & 6.33$\pm$0.05 & \underline{4.85$\pm$0.34} & 6.01$\pm$0.41 &
4.84$\pm$0.17 \textcolor{magenta}{\scriptsize{(-0.21\%)}} & \textbf{4.24$\pm$0.25} \textcolor{magenta}{\scriptsize{(-12.58\%)}}\\
ibm03   & 2.57$\pm$0.07 & 3.08$\pm$0.17 & \underline{1.82$\pm$0.10} & 2.15$\pm$0.17 &
1.75$\pm$0.07 \textcolor{magenta}{\scriptsize{(-3.85\%)}} & \textbf{1.64$\pm$0.06} \textcolor{magenta}{\scriptsize{(-9.89\%)}}\\
ibm04   & 5.73$\pm$0.15 & 4.80$\pm$0.26 & \underline{4.73$\pm$0.07} & 5.00$\pm$0.14 &
4.19$\pm$0.11 \textcolor{magenta}{\scriptsize{(-11.42\%)}} & \textbf{4.06$\pm$0.13} \textcolor{magenta}{\scriptsize{(-14.16\%)}}\\
\bottomrule
\end{tabular}
\end{adjustbox}

\end{small}

\end{center}
\end{table*}

\section{Experiment}
\textbf{Benchmarks and Settings.} 
We evaluate ChiPFormer with the recent advanced RL-based methods  GraphPlace \cite{mirhoseini2021graph}, DeepPR \cite{cheng2021joint}, MaskPlace \cite{lai2022maskplace}, PRNet \cite{chengpolicy}, the supervised learning-based methods Flora \cite{liu2022floorplanning} and GraphPlanner \cite{liu2022graphplanner}, the optimization-based methods DREAMPlace \cite{lin2020dreamplace} and RePlAce \cite{cheng2018replace}, and the manual design approach (Human). All previous methods are implemented by their default settings.  Our experiments are evaluated over 32 circuits from a public placement benchmark (containing  \textit{ISPD05} \cite{nam2005ispd2005}, \textit{ICCAD04} \cite{adya2009iccad}, \textit{Ariane RISC-V CPU} \cite{zaruba2019cost} with a total of 26 circuits) and a private realistic industrial placement task (containing 6 circuits), which involves considerably more chip circuits than reported in previous works. 
More details about the experimental benchmark information and hyperparameter settings can be found in 
Appendix Table~\ref{statistics}, ~\ref{statistics-industrial}, and~\ref{configuration-tab}.

\textbf{Offline Data Collection.} 
To collect our multi-task offline placement data, we employ MaskPlace as the proxy method to learn 12 expert-level behavior (data-collecting) policies over 12 circuits (8 from \textit{ISPD05} and 4 from \textit{ICCAD04}) respectively. 
Then, we collect 500 expert-level placement results for each circuit using these trained data-collecting policies.
Because MaskPlace is based on the stochastic policy,
we can ensure that the collected 500 behaviors will not collapse to 
a single solution for each task, 
thus exhibiting an essential structural diversity in the offline dataset.

\textbf{Macro Placement Results.} 
First, we investigate whether our offline placement formulation can produce effective macro placement results. In this case, we use part of the collected offline data to learn a ChiPFormer placement policy and then finetune the pretrained policy to the unseen circuits. Specifically, we divide the 12 circuits used for the offline data collection into four circuit groups according to the index in the circuit name. 
For example, the 1st group includes the circuits \textit{adaptec1}, \textit{bigblue1} and \textit{ibm01}. 
Then, we use the three grouped offline data to learn a ChiPFormer policy and use the pretrained ChiPFormer as a finetuning initialization for the unseen circuits in the other group. 
We report the best HPWL of macro placement within the limited rollout times over the unseen circuits in Table~\ref{macro-placement}, where all results (mean and standard deviation) are achieved across 5 seeds. 
We can find that ChiPFormer(2k) consistently surpasses the other baselines while improving the placement sample efficiency. We can also observe that even with a substantially small number of rollout times (3k in baselines \textit{v.s.} 300 in ChiPFormer), ChiPFormer(300) can achieve better performance in 10 out of 12 tasks. Due to the superior sample efficiency and the strong performance, we use ChiPFormer(300) as the default finetuning setting in the following experiments when it is not explicitly specified.

Further, we report the overlap ratio and the congestion comparison results in Appendix Table~\ref{overlap} and Fig.\ref{congestion}. Our ChiPFormer consistently outperforms baselines and exhibits better results in both metrics.

\begin{figure*}[!htb]
\centering
\includegraphics[width=2\columnwidth]{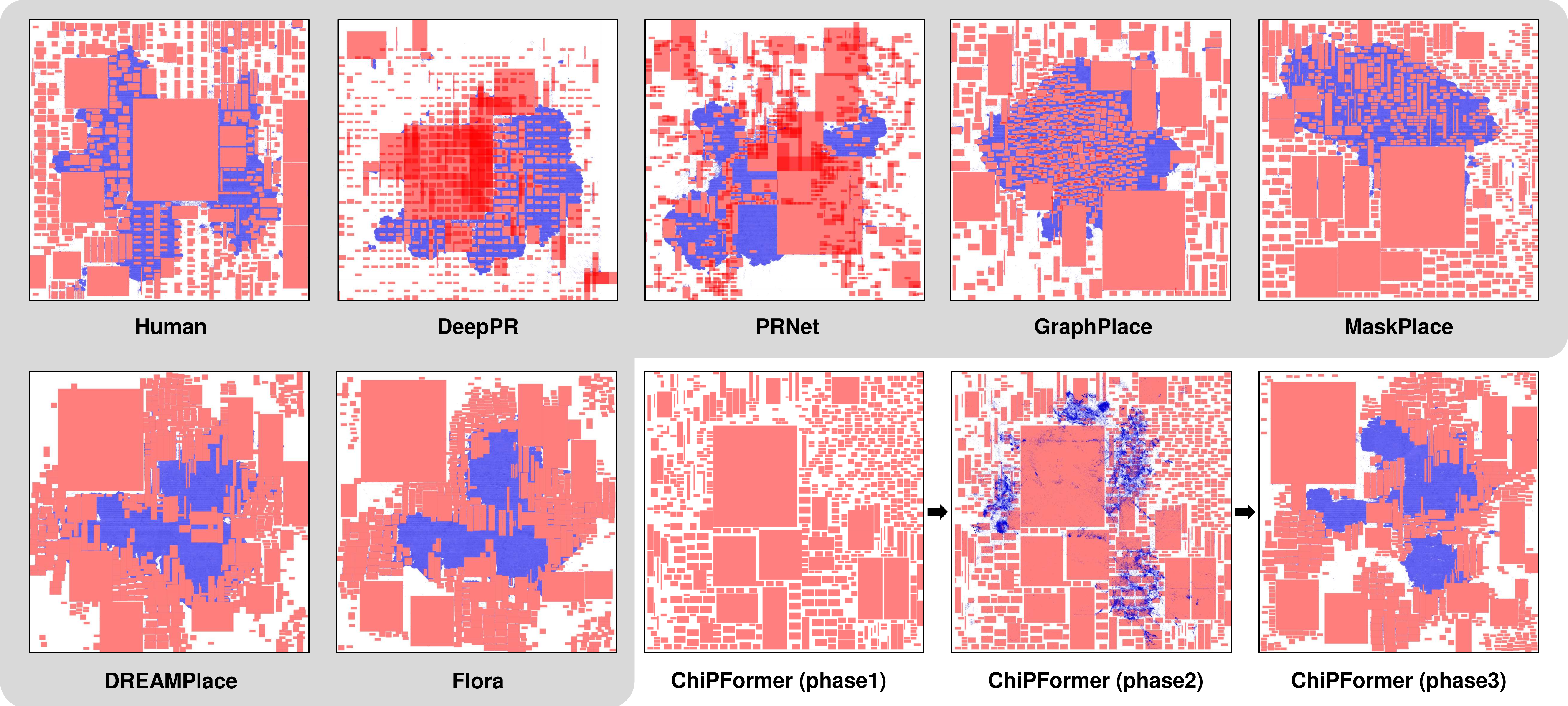}
\caption{\textbf{Visualization of mixed-size placement for circuit \textit{adaptec3}.} \textcolor{red}{Red} marks macros and \textcolor{blue}{blue} marks standard cells. Baseline methods (shaded background) include Human (HPWL = 19.41$\times$$10^7$, Overlap = 0\%), DeepPR (HPWL = 24.11$\times$$10^7$, Overlap = 24.35\%), PRNet (HPWL = 23.24$\times$$10^7$, Overlap = 10.71\%), GraphPlace (HPWL = 25.80$\times$$10^7$, Overlap = 1.24\%), MaskPlace (HPWL = 21.49$\times$$10^7$, Overlap = 0\%), DREAMPlace (HPWL = 15.63$\times$$10^7$, Overlap = 0\%), Flora (HPWL = 15.65$\times$$10^7$, Overlap = 0\%), and ChiPFormer (HPWL = \underline{13.97$\times$$10^7$}, Overlap = \underline{0\%}). ChiPFormer (phase 1), (phase 2), and (phase 3) represent  the return results of phase 1, 2, and 3 in our placement flow (Table \ref{placementflow}), respectively. 
}
\label{visualization}
\end{figure*}

\textbf{Mixed-Size Placement Results.} 
Next, we compare our ChiPFormer to prior baselines (Human, DREAMPlace, GraphPlace, PRNet, DeepPR, MaskPlace, Flora, and GraphPlanner) on the mixed-size placement task. We provide the visualization in Fig.\ref{visualization} and the comparison results in Table \ref{full-placement} and Appendix Table~\ref{full-placement-iccad}. 
Benefiting from the offline pretraining, 
we can observe that our ChiPFormer can produce state-of-the-art placement quality in all mixed-size placements. ChiPFormer can even reduce the HPWL by more than 10\% in several chip placement tasks compared to the best results among the baseline methods.

\begin{table*}[!htb]
\caption{
\textbf{Comparisons of HPWL ($\times$$10^7$) for the mixed-size placement (lower is better).} The baseline results for DREAMPlace, GraphPlace, PRNet, DeepPR, MaskPlace, Flora, and GraphPlanner are taken from the respective papers. Baseline \textit{Human} means the macros are placed by hardware experts. We implement our ChiPFormer method according to the workflow specified in Table \ref{placementflow}. The percentage value in the \textcolor{magenta}{pink} denotes the reduction rate of HPWL compared to the best results (\underline{underlined}) among baselines.
}
\label{full-placement}
\begin{center}
\begin{adjustbox}{max width=2\columnwidth}
\begin{tabular}{ccccccccc|c}
\toprule circuit & Human & DREAMPlace & GraphPlace & PRNet & DeepPR & MaskPlace & Flora & GraphPlanner & ChiPFormer(workflow) \\
\midrule 
adaptec1   & 7.33 & 6.56 & 8.67 & 8.28 & 8.01 & 7.93 & \underline{6.47} & 6.55 & \textbf{6.45$\pm$0.02} \textcolor{magenta}{\scriptsize{(-0.31\%)}} \\
adaptec2   & 8.22 & 10.11 & 12.41 & 12.33 & 12.32 & 9.95 & 7.77 & \underline{7.75} & \textbf{7.36$\pm$0.26} \textcolor{magenta}{\scriptsize{(-5.03\%)}} \\
adaptec3   & 19.41 & 15.63 & 25.80 & 23.24 & 24.11 & 21.49 & 15.65 & \underline{15.08} & \textbf{13.97$\pm$0.80} \textcolor{magenta}{\scriptsize{(-7.36\%)}}\\
adaptec4   & 17.44 & 14.41 & 25.58 & 23.40 & 23.64 & 22.97 & \underline{14.30} & 14.27 & \textbf{12.97$\pm$0.29} \textcolor{magenta}{\scriptsize{(-9.30\%)}}\\
bigblue1   & 8.94 & 8.52 & 16.85 & 14.10 & 14.04 & 9.43 &  \underline{8.51} & 8.59 & \textbf{8.48$\pm$0.02} \textcolor{magenta}{\scriptsize{(-0.35\%)}}\\
bigblue2   & 13.67 & \underline{12.57} & 14.20 & 14.48 & 14.04 & 14.13 & 12.59 & 12.72 & \textbf{9.86$\pm$0.32} \textcolor{magenta}{\scriptsize{(-21.56\%)}}\\
bigblue3 & \underline{30.40} & 46.06& 36.48 & 46.86 & 45.06 & 37.29 & - & - & \textbf{27.33$\pm$0.31} \textcolor{magenta}{\scriptsize{(-10.10\%)}}\\
bigblue4 & \underline{74.38} & 79.50 & 104.00 & 100.13 & 95.20 & 106.18  & 74.76 & - &  \textbf{65.98$\pm$4.08} \textcolor{magenta}{\scriptsize{(-11.29\%)}}\\
\bottomrule
\end{tabular}
\end{adjustbox}
\end{center}
\end{table*}

\begin{figure}[!htb]
\begin{center}
\centerline{\includegraphics[width=\columnwidth]{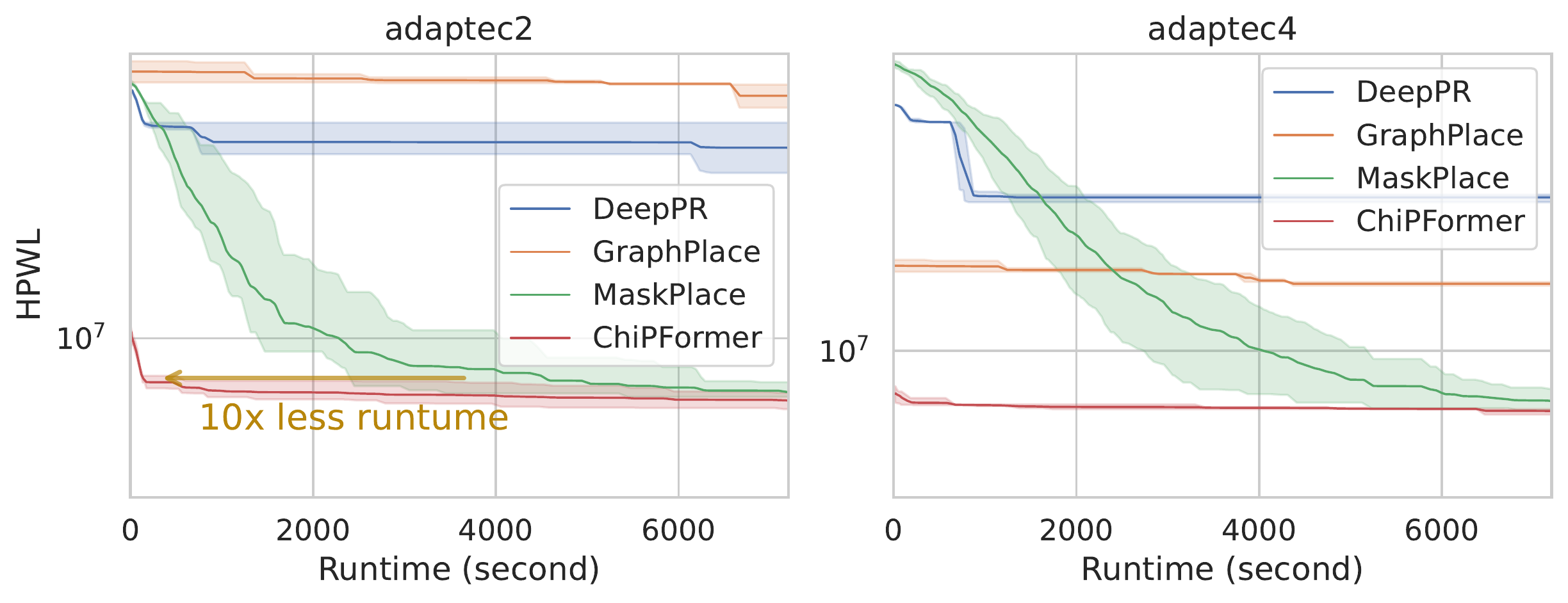}}
\caption{
\textbf{HPWL curves over runtime.} All methods are evaluated on two RTX3090 GPUs.  ChiPFormer can search for high-quality placements in a very short time, more than 10 times faster compared with other methods.
}
\label{timeline}
\end{center}
\end{figure}

\textbf{Time Efficiency.}
Beyond the training sample efficiency explored above, we investigate the training time efficiency. We provide plots of the HPWL metric over runtime in Fig.\ref{timeline}. Our offline ChiPFormer finetuning is consistently competitive with previous online methods with more stable performance and less running time, leading to more than $10\times$ speed-up in time efficiency.

\textbf{Industrial Chip Placement.} In Table \ref{real-world}, we also compare ChiPFormer to the human experts in 6 realistic industrial chip design tasks. Compared to the public placement benchmarks, realistic industrial placement involves more complex constraints and design metrics \citep{bhasker2009static, wang2009electronic}. 
For a fair comparison, we use industrial EDA tool 
to conduct the mixed-size placement workflow after applying our pretrained ChiPFormer model.
We provide the comparison results in Table \ref{real-world}, where we can find that ChiPFormer can exceed expert human performance in most industrial metrics. 

\begin{table}[!htb]
\caption{
\textbf{Comparisons to human expert design on 6 industrial chip design tasks.} The lower the metrics the better.
}
\label{real-world}
    \centering
    \begin{adjustbox}{max width=0.9\columnwidth}
    \begin{threeparttable}
    
    \begin{tabular}{cccccccc}
    \toprule
        circuit & method & \multicolumn{2}{c}{Timing} & NVP\tnote{1} & \multicolumn{2}{c}{Congestion}   \\ 
    \midrule
        ~ & ~ & WNS/ps & TNS/ns & ~ & H/\% & V/\%  \\ \midrule
    
        \multirow{3}{*}{C1} & Human & 204 & 57.1 & 2569 & 0.06 & 0.38  \\  
       ~ & MaskPlace & 161 & 42.7 & 1964 & 0.07 & \textbf{0.07} \\
        ~ & ChiPFormer & \textbf{142} & \textbf{19.4} & \textbf{1636} & \textbf{0.04} & \textbf{0.07}  \\  \midrule
         \multirow{3}{*}{C2} & Human & 403 & 492.2 & 11360 & 0.63 & 2.05  \\ 
         ~ & MaskPlace & 242 & 259.1 & 9710 & 0.57 & 1.67 \\
        ~ & ChiPFormer & \textbf{177} & \textbf{224.9} & \textbf{8110} & \textbf{0.53} & \textbf{1.27}  \\ \midrule
         \multirow{3}{*}{C3} & Human & \textbf{102} & 91.9 & 5614 & \textbf{1.02} & 0.85  \\ 
         ~ & MaskPlace & 116 & 92.8 & 5559 & 1.05 & 0.87 \\
        ~ & ChiPFormer & 108 & \textbf{91.2} & \textbf{5452} & \textbf{1.02} & \textbf{0.82}  \\ \midrule
         \multirow{3}{*}{C4} & Human & 399 & 438.0 & 13925 & 0.97 & \textbf{0.34}  \\ 
         ~ & MaskPlace & 389 & 324.2 &12582 & 0.68 & \textbf{0.34}\\
        ~ & ChiPFormer & \textbf{248} & \textbf{266.0} & \textbf{12398} & \textbf{0.62} & \textbf{0.34}  \\ \midrule
        \multirow{3}{*}{C5} & Human & 89 & 10.8 & 2675 & \textbf{0.02} & 0.07 \\
        ~ & MaskPlace & 122 & 32.2 & 2975& \textbf{0.02} & 0.22 \\
        ~ & ChiPFormer & \textbf{80} & \textbf{4.9} & \textbf{1706} & \textbf{0.02} & \textbf{0.04} \\  \midrule
        \multirow{3}{*}{C6} & Human & 154 & 137.4 & 6833 & 0.70 & 0.22 \\
        ~ & MaskPlace & 81 & 49.6& 7040 & 0.77 & 0.26\\
        ~ & ChiPFormer & \textbf{78} & \textbf{38.1} & \textbf{6412} & \textbf{0.63} & \textbf{0.22} \\ 
    \bottomrule
    \end{tabular}
    \begin{tablenotes}
    \tiny
    \item[1] NVP = Number of Violation Points, WNS = Worst Negative Slack, TNS = Total Negative Slack.\\
    More details about metrics can be found in \citet{bhasker2009static} and \citet{wang2009electronic}.
    \end{tablenotes}
    \end{threeparttable}
    \end{adjustbox}
\end{table}

\begin{figure}[!htb]
\centering
     \begin{subfigure}[b]{0.49\columnwidth}
         \centering
         \includegraphics[width=\columnwidth]{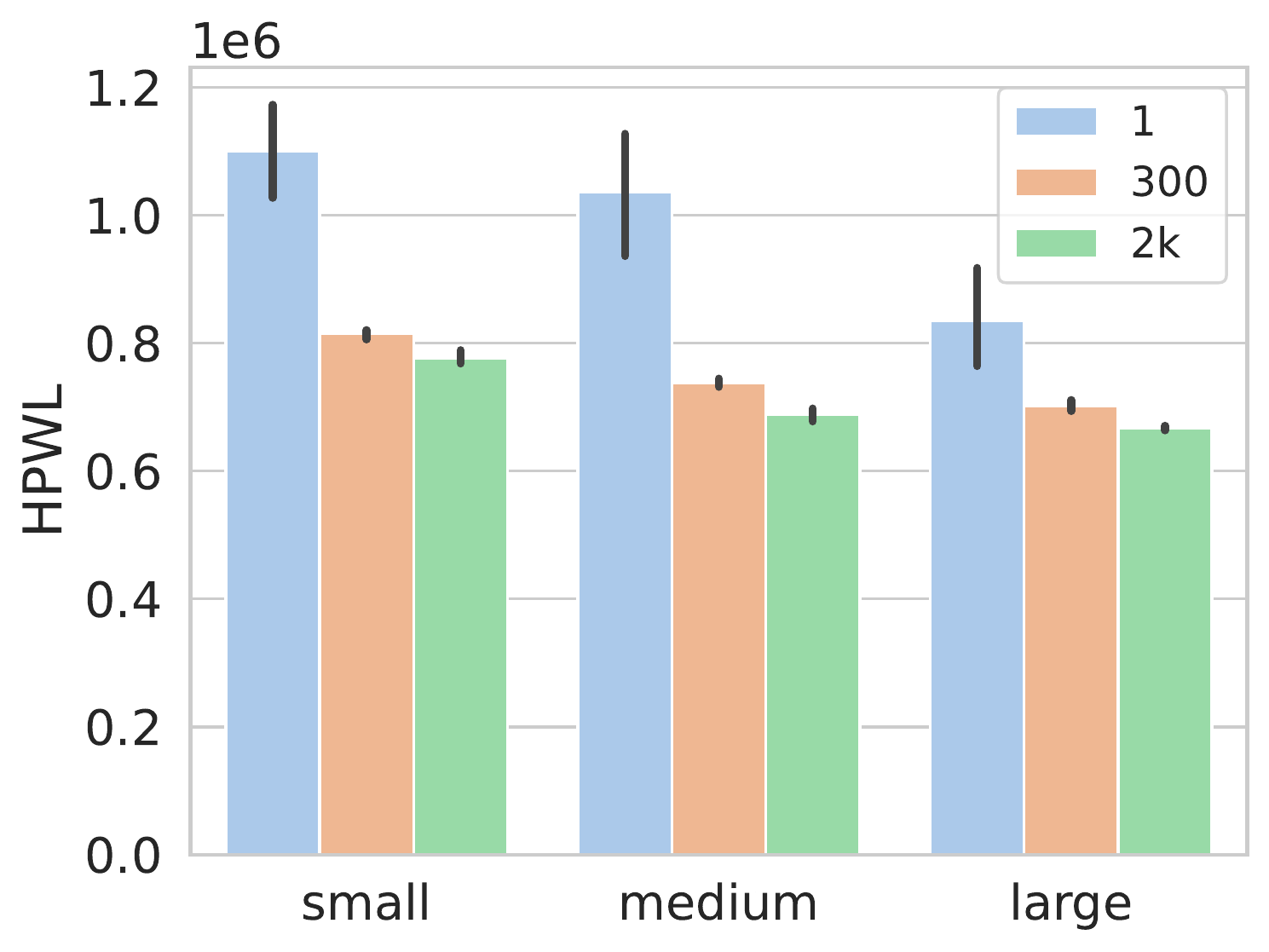}
         \caption{}
         \label{fig:y equals x}
     \end{subfigure}
     \hfill
     \begin{subfigure}[b]{0.49\columnwidth}
         \centering
         \includegraphics[width=\columnwidth]{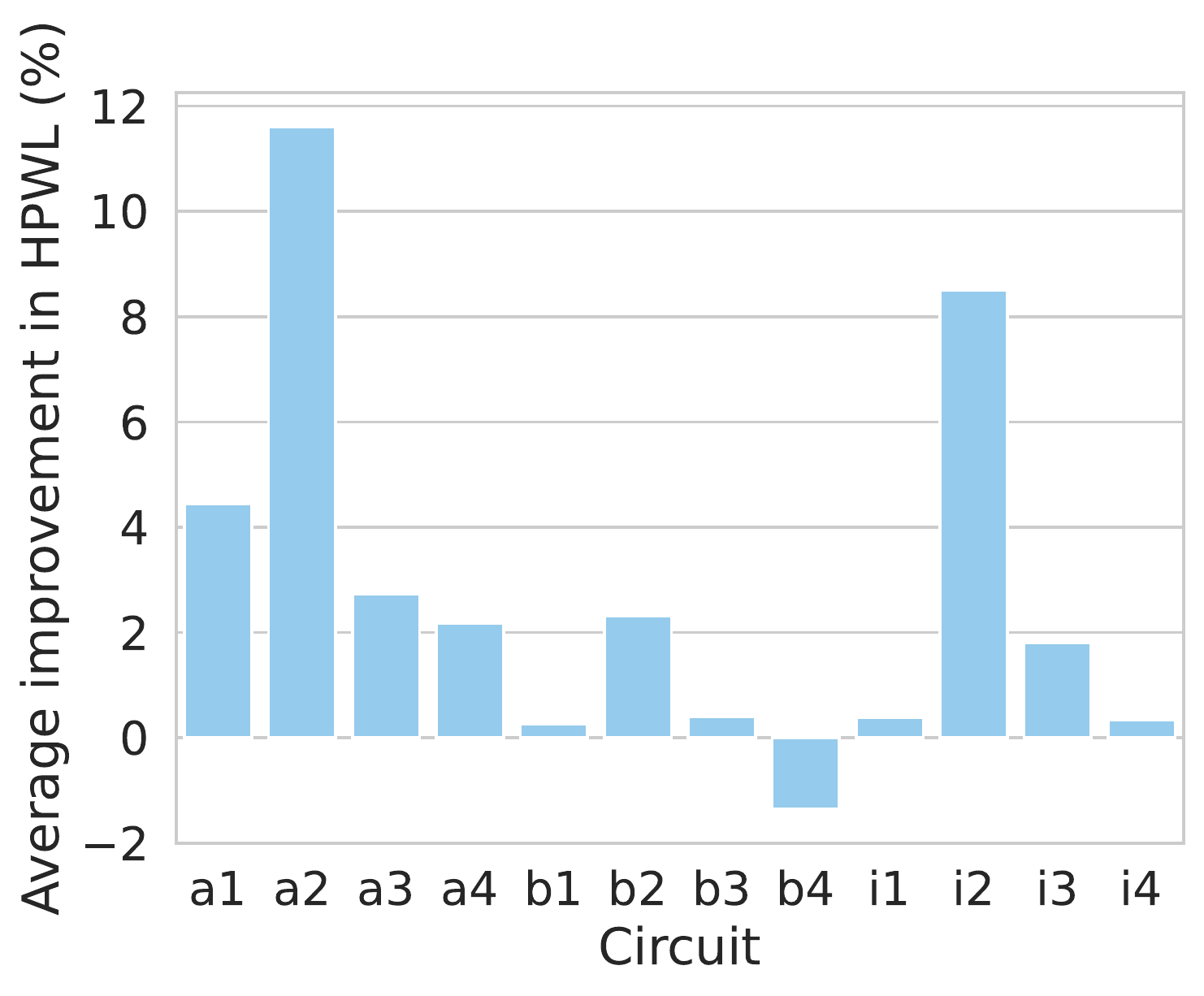}
         \caption{}
         \label{circuit-token-ablation}
     \end{subfigure}
     \caption{
     \textbf{(a) Comparison of different offline dataset sizes.} Small, medium and large datasets include 1/3/9 offline circuits, respectively. We can see that the larger dataset tends to generate higher-quality placements.  
    \textbf{(b) Performance improvement when introducing the circuit token.} We can observe that introducing the circuit token contributes to improved performance in most circuits.
    }
    \label{dataset-circuit-token}
\end{figure}

\textbf{Dataset Size.} 
To explore the impact of training dataset size on the transfer ability of our ChiPFormer, we construct three offline datasets (small, medium, and large) containing 1, 3, and 9 circuits, respectively.
For each dataset, we compare the HPWL results of ChiPFormer(1), ChiPFormer(300), and ChiPFormer(2k) after finetuning for circuit \textit{adaptec1} over 10 seeds in Fig.\ref{dataset-circuit-token} (a). The results show that using more offline circuits, ChiPFormer can yield better generalization for new unseen circuits, especially in the zero-shot transfer settings (ChiPFormer(1)).

\textbf{Ablation Study.} To study the role of the multi-task hindsight information matching and verify the effects of circuit token $\text{HI}(\bc)$, we ablate the circuit token and keep other settings the same. 
We test the zero-shot performance on 12 circuits in benchmark \textit{ISPD05} and \textit{ICCAD04}. In Fig.\ref{dataset-circuit-token} (b), we measure the difference between the ablated ChiPFormer and the full ChiPFormer. 
We can find that when introducing the circuit token, ChiPFormer can acquire a better generalization ability to zero-shot to unseen tasks.

\begin{figure}[!htb]
\begin{center}
\centerline{\includegraphics[width=1.0\columnwidth]{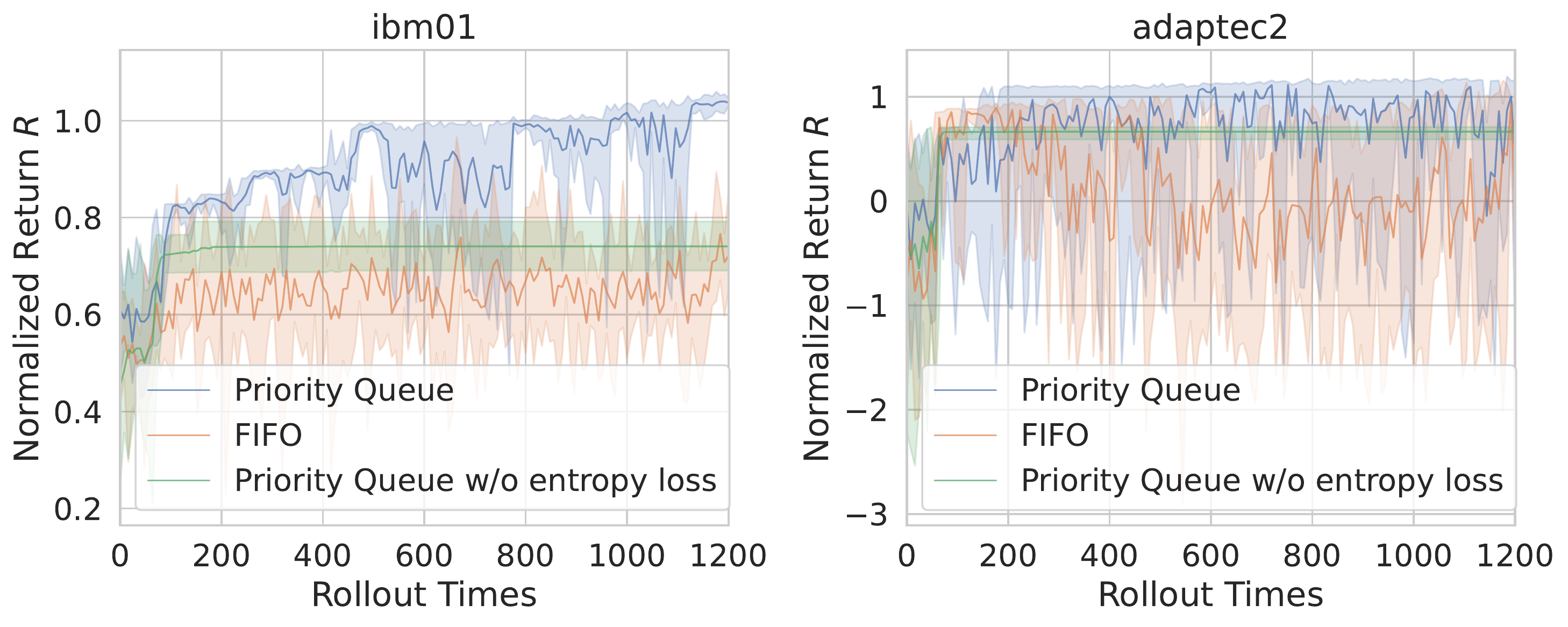}}
\caption{
\textbf{Ablation results in replay buffer update rules and the entropy motivation in ChiPFormer finetuning.}}
\label{fifo-priority}
\end{center}
\vspace{-22pt}
\end{figure}

Further, we compare FIFO and priority-based storage strategies for the finetuning replay buffer and ablate the entropy loss in Eqn.\eqref{eq:finetuning}. 
We show the ablation results in Fig.\ref{fifo-priority}. We can see that the FIFO-based strategy tends to forget the experiences with higher returns and lead to sub-optimal and high-variance placement results. When removing the entropy loss, ChiPFormer suffers from a lack of exploration ability and converges to a local optimum. 

\section{Conclusion}
In this paper, we propose ChiPFormer, an offline RL method for chip placement tasks, and design a mixed-size placement workflow. ChiPFormer can learn from existing placement solutions (offline data) and significantly improve the training sample and time efficiency. ChiPFormer can also acquire strong zero-shot transfer ability and yield an effective initialization for the few-shot finetuning on unseen chip placement~tasks.

\section*{Acknowledgements}
This paper is partially supported by the National Key R\&D Program of China No.2022ZD0161000 and the General Research Fund of Hong Kong No.17200622.
\bibliography{example_paper}

\begin{thebibliography}{67}
\providecommand{\natexlab}[1]{#1}
\providecommand{\url}[1]{\texttt{#1}}
\expandafter\ifx\csname urlstyle\endcsname\relax
  \providecommand{\doi}[1]{doi: #1}\else
  \providecommand{\doi}{doi: \begingroup \urlstyle{rm}\Url}\fi

\bibitem[Adya et~al.(2009)Adya, Chaturvedi, and Markov]{adya2009iccad}
Adya, S., Chaturvedi, S., and Markov, I.
\newblock Iccad’04 mixed-size placement benchmarks.
\newblock \emph{GSRC Bookshelf}, 2009.

\bibitem[Bhasker \& Chadha(2009)Bhasker and Chadha]{bhasker2009static}
Bhasker, J. and Chadha, R.
\newblock \emph{Static timing analysis for nanometer designs: A practical
  approach}.
\newblock Springer Science \& Business Media, 2009.

\bibitem[Brenner et~al.(2015)Brenner, Hermann, Hoppmann, and
  Ochsendorf]{brenner2015bonnplace}
Brenner, U., Hermann, A., Hoppmann, N., and Ochsendorf, P.
\newblock Bonnplace: A self-stabilizing placement framework.
\newblock In \emph{Proceedings of the 2015 Symposium on International Symposium
  on Physical Design (ISPD)}, pp.\  9--16, 2015.

\bibitem[Chan et~al.(2006)Chan, Cong, Shinnerl, Sze, and Xie]{chan2006mpl6}
Chan, T.~F., Cong, J., Shinnerl, J.~R., Sze, K., and Xie, M.
\newblock mpl6: Enhanced multilevel mixed-size placement.
\newblock In \emph{Proceedings of the 2006 international symposium on Physical
  design (ISPD)}, pp.\  212--214, 2006.

\bibitem[Chebotar et~al.(2021)Chebotar, Hausman, Lu, Xiao, Kalashnikov, Varley,
  Irpan, Eysenbach, Julian, Finn, et~al.]{chebotar2021actionable}
Chebotar, Y., Hausman, K., Lu, Y., Xiao, T., Kalashnikov, D., Varley, J.,
  Irpan, A., Eysenbach, B., Julian, R., Finn, C., et~al.
\newblock Actionable models: Unsupervised offline reinforcement learning of
  robotic skills.
\newblock \emph{arXiv preprint arXiv:2104.07749}, 2021.

\bibitem[Chen et~al.(2021)Chen, Lu, Rajeswaran, Lee, Grover, Laskin, Abbeel,
  Srinivas, and Mordatch]{chen2021decision}
Chen, L., Lu, K., Rajeswaran, A., Lee, K., Grover, A., Laskin, M., Abbeel, P.,
  Srinivas, A., and Mordatch, I.
\newblock Decision transformer: Reinforcement learning via sequence modeling.
\newblock \emph{Advances in neural information processing systems (NeurIPS)},
  34:\penalty0 15084--15097, 2021.

\bibitem[Chen et~al.(2008)Chen, Jiang, Hsu, Chen, and Chang]{chen2008ntuplace3}
Chen, T.-C., Jiang, Z.-W., Hsu, T.-C., Chen, H.-C., and Chang, Y.-W.
\newblock Ntuplace3: An analytical placer for large-scale mixed-size designs
  with preplaced blocks and density constraints.
\newblock \emph{IEEE Transactions on Computer-Aided Design of Integrated
  Circuits and Systems (TCAD)}, 27\penalty0 (7):\penalty0 1228--1240, 2008.

\bibitem[Cheng et~al.(2018)Cheng, Kahng, Kang, and Wang]{cheng2018replace}
Cheng, C.-K., Kahng, A.~B., Kang, I., and Wang, L.
\newblock Replace: Advancing solution quality and routability validation in
  global placement.
\newblock \emph{IEEE Transactions on Computer-Aided Design of Integrated
  Circuits and Systems (TCAD)}, 38\penalty0 (9):\penalty0 1717--1730, 2018.

\bibitem[Cheng \& Yan(2021)Cheng and Yan]{cheng2021joint}
Cheng, R. and Yan, J.
\newblock On joint learning for solving placement and routing in chip design.
\newblock \emph{Advances in Neural Information Processing Systems (NeurIPS)},
  34, 2021.

\bibitem[Cheng et~al.(2022)Cheng, Lyu, Li, Ye, Jianye, and Yan]{chengpolicy}
Cheng, R., Lyu, X., Li, Y., Ye, J., Jianye, H., and Yan, J.
\newblock The policy-gradient placement and generative routing neural networks
  for chip design.
\newblock In \emph{Advances in Neural Information Processing Systems
  (NeurIPS)}, 2022.

\bibitem[Davies et~al.(2021)Davies, Veli{\v{c}}kovi{\'c}, Buesing, Blackwell,
  Zheng, Toma{\v{s}}ev, Tanburn, Battaglia, Blundell, Juh{\'a}sz,
  et~al.]{davies2021advancing}
Davies, A., Veli{\v{c}}kovi{\'c}, P., Buesing, L., Blackwell, S., Zheng, D.,
  Toma{\v{s}}ev, N., Tanburn, R., Battaglia, P., Blundell, C., Juh{\'a}sz, A.,
  et~al.
\newblock Advancing mathematics by guiding human intuition with ai.
\newblock \emph{Nature}, 600\penalty0 (7887):\penalty0 70--74, 2021.

\bibitem[Emmons et~al.(2021)Emmons, Eysenbach, Kostrikov, and
  Levine]{emmons2021rvs}
Emmons, S., Eysenbach, B., Kostrikov, I., and Levine, S.
\newblock Rvs: What is essential for offline rl via supervised learning?
\newblock \emph{arXiv preprint arXiv:2112.10751}, 2021.

\bibitem[Fu et~al.(2020)Fu, Kumar, Nachum, Tucker, and Levine]{fu2020d4rl}
Fu, J., Kumar, A., Nachum, O., Tucker, G., and Levine, S.
\newblock D4rl: Datasets for deep data-driven reinforcement learning.
\newblock \emph{arXiv preprint arXiv:2004.07219}, 2020.

\bibitem[Fujimoto et~al.(2019)Fujimoto, Meger, and Precup]{fujimoto2019off}
Fujimoto, S., Meger, D., and Precup, D.
\newblock Off-policy deep reinforcement learning without exploration.
\newblock In \emph{International conference on machine learning (ICML)}, pp.\
  2052--2062. PMLR, 2019.

\bibitem[Furuta et~al.(2021)Furuta, Matsuo, and Gu]{furuta2021generalized}
Furuta, H., Matsuo, Y., and Gu, S.~S.
\newblock Generalized decision transformer for offline hindsight information
  matching.
\newblock \emph{arXiv preprint arXiv:2111.10364}, 2021.

\bibitem[Garey \& Johnson(1977)Garey and Johnson]{garey1977rectilinear}
Garey, M.~R. and Johnson, D.~S.
\newblock The rectilinear steiner tree problem is np-complete.
\newblock \emph{SIAM Journal on Applied Mathematics}, 32\penalty0 (4):\penalty0
  826--834, 1977.

\bibitem[Ghosh et~al.(2019)Ghosh, Gupta, Reddy, Fu, Devin, Eysenbach, and
  Levine]{ghosh2019learning}
Ghosh, D., Gupta, A., Reddy, A., Fu, J., Devin, C., Eysenbach, B., and Levine,
  S.
\newblock Learning to reach goals via iterated supervised learning.
\newblock \emph{arXiv preprint arXiv:1912.06088}, 2019.

\bibitem[Gu et~al.(2020)Gu, Jiang, Lin, and Pan]{gu2020dreamplace}
Gu, J., Jiang, Z., Lin, Y., and Pan, D.~Z.
\newblock Dreamplace 3.0: Multi-electrostatics based robust vlsi placement with
  region constraints.
\newblock In \emph{2020 IEEE/ACM International Conference On Computer Aided
  Design (ICCAD)}, pp.\  1--9. IEEE, 2020.

\bibitem[Jiang et~al.(2021)Jiang, Songhori, Wang, Goldie, Mirhoseini, Jiang,
  Lee, and Pan]{jiang2021delving}
Jiang, Z., Songhori, E., Wang, S., Goldie, A., Mirhoseini, A., Jiang, J., Lee,
  Y.-J., and Pan, D.~Z.
\newblock Delving into macro placement with reinforcement learning.
\newblock In \emph{2021 ACM/IEEE 3rd Workshop on Machine Learning for CAD
  (MLCAD)}, pp.\  1--3. IEEE, 2021.

\bibitem[Jumper et~al.(2021)Jumper, Evans, Pritzel, Green, Figurnov,
  Ronneberger, Tunyasuvunakool, Bates, {\v{Z}}{\'\i}dek, Potapenko,
  et~al.]{jumper2021highly}
Jumper, J., Evans, R., Pritzel, A., Green, T., Figurnov, M., Ronneberger, O.,
  Tunyasuvunakool, K., Bates, R., {\v{Z}}{\'\i}dek, A., Potapenko, A., et~al.
\newblock Highly accurate protein structure prediction with alphafold.
\newblock \emph{Nature}, 596\penalty0 (7873):\penalty0 583--589, 2021.

\bibitem[Kahng et~al.(2005)Kahng, Reda, and Wang]{kahng2005aplace}
Kahng, A.~B., Reda, S., and Wang, Q.
\newblock Aplace: A general analytic placement framework.
\newblock In \emph{Proceedings of the 2005 international symposium on Physical
  design (ISPD)}, pp.\  233--235, 2005.

\bibitem[Kang et~al.(2023)Kang, Shi, Liu, He, and Wang]{kang2023beyond}
Kang, Y., Shi, D., Liu, J., He, L., and Wang, D.
\newblock Beyond reward: Offline preference-guided policy optimization.
\newblock \emph{arXiv preprint arXiv:2305.16217}, 2023.

\bibitem[Khatkhate et~al.(2004)Khatkhate, Li, Agnihotri, Yildiz, Ono, Koh, and
  Madden]{khatkhate2004recursive}
Khatkhate, A., Li, C., Agnihotri, A.~R., Yildiz, M.~C., Ono, S., Koh, C.-K.,
  and Madden, P.~H.
\newblock Recursive bisection based mixed block placement.
\newblock In \emph{Proceedings of the 2004 international symposium on Physical
  design (ISPD)}, pp.\  84--89, 2004.

\bibitem[Kim et~al.()Kim, Kim, Kim, and Park]{kimcollaborative}
Kim, H., Kim, M., Kim, J., and Park, J.
\newblock Collaborative symmetricity exploitation for offline learning of
  hardware design solver.
\newblock In \emph{3rd Offline RL Workshop: Offline RL as a``Launchpad''}.

\bibitem[Kim \& Markov(2012)Kim and Markov]{kim2012complx}
Kim, M.-C. and Markov, I.~L.
\newblock Complx: A competitive primal-dual lagrange optimization for global
  placement.
\newblock In \emph{Proceedings of the 49th Annual Design Automation Conference
  (DAC)}, pp.\  747--752, 2012.

\bibitem[Kim et~al.(2012)Kim, Viswanathan, Alpert, Markov, and
  Ramji]{kim2012maple}
Kim, M.-C., Viswanathan, N., Alpert, C.~J., Markov, I.~L., and Ramji, S.
\newblock Maple: Multilevel adaptive placement for mixed-size designs.
\newblock In \emph{Proceedings of the 2012 ACM international symposium on
  International Symposium on Physical Design (ISPD)}, pp.\  193--200, 2012.

\bibitem[Kipf \& Welling(2016{\natexlab{a}})Kipf and Welling]{kipf2016semi}
Kipf, T.~N. and Welling, M.
\newblock Semi-supervised classification with graph convolutional networks.
\newblock \emph{arXiv preprint arXiv:1609.02907}, 2016{\natexlab{a}}.

\bibitem[Kipf \& Welling(2016{\natexlab{b}})Kipf and
  Welling]{kipf2016variational}
Kipf, T.~N. and Welling, M.
\newblock Variational graph auto-encoders.
\newblock \emph{arXiv preprint arXiv:1611.07308}, 2016{\natexlab{b}}.

\bibitem[Lai et~al.(2020)Lai, Ping, Wu, Lu, and Ye]{lai2020opensmax}
Lai, Y., Ping, G., Wu, Y., Lu, C., and Ye, X.
\newblock Opensmax: Unknown domain generation algorithm detection.
\newblock In \emph{24th European Conference on Artificial Intelligence (ECAI)},
  pp.\  1850--1857. IOS Press, 2020.

\bibitem[Lai et~al.(2022)Lai, Mu, and Luo]{lai2022maskplace}
Lai, Y., Mu, Y., and Luo, P.
\newblock Maskplace: Fast chip placement via reinforced visual representation
  learning.
\newblock In \emph{Advances in Neural Information Processing Systems
  (NeurIPS)}, 2022.

\bibitem[Lee et~al.(2022)Lee, Seo, Lee, Abbeel, and Shin]{lee2022offline}
Lee, S., Seo, Y., Lee, K., Abbeel, P., and Shin, J.
\newblock Offline-to-online reinforcement learning via balanced replay and
  pessimistic q-ensemble.
\newblock In \emph{Conference on Robot Learning (CoRL)}, pp.\  1702--1712.
  PMLR, 2022.

\bibitem[Levine et~al.(2020)Levine, Kumar, Tucker, and Fu]{levine2020offline}
Levine, S., Kumar, A., Tucker, G., and Fu, J.
\newblock Offline reinforcement learning: Tutorial, review, and perspectives on
  open problems.
\newblock \emph{arXiv preprint arXiv:2005.01643}, 2020.

\bibitem[Lin et~al.(2013)Lin, Chu, Shinnerl, Bustany, and
  Nedelchev]{lin2013polar}
Lin, T., Chu, C., Shinnerl, J.~R., Bustany, I., and Nedelchev, I.
\newblock Polar: Placement based on novel rough legalization and refinement.
\newblock In \emph{2013 IEEE/ACM International Conference on Computer-Aided
  Design (ICCAD)}, pp.\  357--362. IEEE, 2013.

\bibitem[Lin et~al.(2020)Lin, Jiang, Gu, Li, Dhar, Ren, Khailany, and
  Pan]{lin2020dreamplace}
Lin, Y., Jiang, Z., Gu, J., Li, W., Dhar, S., Ren, H., Khailany, B., and Pan,
  D.~Z.
\newblock Dreamplace: Deep learning toolkit-enabled gpu acceleration for modern
  vlsi placement.
\newblock \emph{IEEE Transactions on Computer-Aided Design of Integrated
  Circuits and Systems (TCAD)}, 40\penalty0 (4):\penalty0 748--761, 2020.

\bibitem[Liu et~al.(2021)Liu, Shen, Wang, Kang, and Tian]{liu2021unsupervised}
Liu, J., Shen, H., Wang, D., Kang, Y., and Tian, Q.
\newblock Unsupervised domain adaptation with dynamics-aware rewards in
  reinforcement learning.
\newblock \emph{Advances in Neural Information Processing Systems (NeurIPS)},
  34:\penalty0 28784--28797, 2021.

\bibitem[Liu et~al.(2022{\natexlab{a}})Liu, Wang, Tian, and Chen]{liu2022learn}
Liu, J., Wang, D., Tian, Q., and Chen, Z.
\newblock Learn goal-conditioned policy with intrinsic motivation for deep
  reinforcement learning.
\newblock In \emph{Proceedings of the AAAI Conference on Artificial
  Intelligence (AAAI)}, volume~36, pp.\  7558--7566, 2022{\natexlab{a}}.

\bibitem[Liu et~al.(2022{\natexlab{b}})Liu, Zhang, and Wang]{liudara2022}
Liu, J., Zhang, H., and Wang, D.
\newblock {DARA:} dynamics-aware reward augmentation in offline reinforcement
  learning.
\newblock In \emph{The Tenth International Conference on Learning
  Representations (ICLR)}. OpenReview.net, 2022{\natexlab{b}}.

\bibitem[Liu et~al.(2022{\natexlab{c}})Liu, Ju, Li, Dong, Zhou, Wang, Yang,
  Zeng, and Shang]{liu2022floorplanning}
Liu, Y., Ju, Z., Li, Z., Dong, M., Zhou, H., Wang, J., Yang, F., Zeng, X., and
  Shang, L.
\newblock Floorplanning with graph attention.
\newblock In \emph{Proceedings of the 59th ACM/IEEE Design Automation
  Conference (DAC)}, pp.\  1303--1308, 2022{\natexlab{c}}.

\bibitem[Liu et~al.(2022{\natexlab{d}})Liu, Ju, Li, Dong, Zhou, Wang, Yang,
  Zeng, and Shang]{liu2022graphplanner}
Liu, Y., Ju, Z., Li, Z., Dong, M., Zhou, H., Wang, J., Yang, F., Zeng, X., and
  Shang, L.
\newblock Graphplanner: Floorplanning with graph neural network.
\newblock \emph{ACM Transactions on Design Automation of Electronic Systems
  (TODAES)}, 2022{\natexlab{d}}.

\bibitem[Lu et~al.(2014)Lu, Chen, Chang, Sha, Dennis, Huang, Teng, and
  Cheng]{lu2014eplace}
Lu, J., Chen, P., Chang, C.-C., Sha, L., Dennis, J., Huang, H., Teng, C.-C.,
  and Cheng, C.-K.
\newblock eplace: Electrostatics based placement using nesterov's method.
\newblock In \emph{2014 51st ACM/EDAC/IEEE Design Automation Conference (DAC)},
  pp.\  1--6. IEEE, 2014.

\bibitem[Lu et~al.(2022)Lu, Yang, Lim, and Ren]{lu2022placement}
Lu, Y.-C., Yang, T., Lim, S.~K., and Ren, H.
\newblock Placement optimization via ppa-directed graph clustering.
\newblock In \emph{2022 ACM/IEEE 4th Workshop on Machine Learning for CAD
  (MLCAD)}, pp.\  1--6. IEEE, 2022.

\bibitem[Mirhoseini et~al.(2021)Mirhoseini, Goldie, Yazgan, Jiang, Songhori,
  Wang, Lee, Johnson, Pathak, Nazi, et~al.]{mirhoseini2021graph}
Mirhoseini, A., Goldie, A., Yazgan, M., Jiang, J.~W., Songhori, E., Wang, S.,
  Lee, Y.-J., Johnson, E., Pathak, O., Nazi, A., et~al.
\newblock A graph placement methodology for fast chip design.
\newblock \emph{Nature}, 594\penalty0 (7862):\penalty0 207--212, 2021.

\bibitem[Nam et~al.(2005)Nam, Alpert, Villarrubia, Winter, and
  Yildiz]{nam2005ispd2005}
Nam, G.-J., Alpert, C.~J., Villarrubia, P., Winter, B., and Yildiz, M.
\newblock The ispd2005 placement contest and benchmark suite.
\newblock In \emph{Proceedings of the 2005 international symposium on Physical
  design (ISPD)}, pp.\  216--220, 2005.

\bibitem[Nocedal \& Wright(2006)Nocedal and Wright]{nocedal2006quadratic}
Nocedal, J. and Wright, S.~J.
\newblock Quadratic programming.
\newblock \emph{Numerical optimization}, pp.\  448--492, 2006.

\bibitem[Radford et~al.(2018)Radford, Narasimhan, Salimans, Sutskever,
  et~al.]{radford2018improving}
Radford, A., Narasimhan, K., Salimans, T., Sutskever, I., et~al.
\newblock Improving language understanding by generative pre-training.
\newblock 2018.

\bibitem[Radford et~al.(2019)Radford, Wu, Child, Luan, Amodei, Sutskever,
  et~al.]{radford2019language}
Radford, A., Wu, J., Child, R., Luan, D., Amodei, D., Sutskever, I., et~al.
\newblock Language models are unsupervised multitask learners.
\newblock \emph{OpenAI blog}, 1\penalty0 (8):\penalty0 9, 2019.

\bibitem[Reed et~al.(2022)Reed, Zolna, Parisotto, Colmenarejo, Novikov,
  Barth-Maron, Gimenez, Sulsky, Kay, Springenberg, et~al.]{reed2022generalist}
Reed, S., Zolna, K., Parisotto, E., Colmenarejo, S.~G., Novikov, A.,
  Barth-Maron, G., Gimenez, M., Sulsky, Y., Kay, J., Springenberg, J.~T.,
  et~al.
\newblock s.
\newblock \emph{arXiv preprint arXiv:2205.06175}, 2022.

\bibitem[Roy et~al.(2006)Roy, Adya, Papa, and Markov]{roy2006min}
Roy, J.~A., Adya, S.~N., Papa, D.~A., and Markov, I.~L.
\newblock Min-cut floorplacement.
\newblock \emph{IEEE Transactions on Computer-Aided Design of Integrated
  Circuits and Systems (TCAD)}, 25\penalty0 (7):\penalty0 1313--1326, 2006.

\bibitem[Schulman et~al.(2017)Schulman, Wolski, Dhariwal, Radford, and
  Klimov]{schulman2017proximal}
Schulman, J., Wolski, F., Dhariwal, P., Radford, A., and Klimov, O.
\newblock Proximal policy optimization algorithms.
\newblock \emph{arXiv preprint arXiv:1707.06347}, 2017.

\bibitem[Shin et~al.(2021)Shin, Brown, and Dragan]{shin2021offline}
Shin, D., Brown, D.~S., and Dragan, A.~D.
\newblock Offline preference-based apprenticeship learning.
\newblock \emph{arXiv preprint arXiv:2107.09251}, 2021.

\bibitem[Spindler et~al.(2008)Spindler, Schlichtmann, and
  Johannes]{spindler2008kraftwerk2}
Spindler, P., Schlichtmann, U., and Johannes, F.~M.
\newblock Kraftwerk2—a fast force-directed quadratic placement approach using
  an accurate net model.
\newblock \emph{IEEE Transactions on Computer-Aided Design of Integrated
  Circuits and Systems (TCAD)}, 27\penalty0 (8):\penalty0 1398--1411, 2008.

\bibitem[Sutton et~al.(1999)Sutton, McAllester, Singh, and
  Mansour]{sutton1999policy}
Sutton, R.~S., McAllester, D., Singh, S., and Mansour, Y.
\newblock Policy gradient methods for reinforcement learning with function
  approximation.
\newblock \emph{Advances in neural information processing systems (NeurIPS)},
  12, 1999.

\bibitem[Tian et~al.(2021{\natexlab{a}})Tian, Liu, Wang, and
  Wang]{tian2021unsupervised}
Tian, Q., Liu, J., Wang, G., and Wang, D.
\newblock Unsupervised discovery of transitional skills for deep reinforcement
  learning.
\newblock In \emph{International Joint Conference on Neural Networks (IJCNN)},
  pp.\  1--8. IEEE, 2021{\natexlab{a}}.

\bibitem[Tian et~al.(2021{\natexlab{b}})Tian, Wang, Liu, Wang, and
  Kang]{tian2021independent}
Tian, Q., Wang, G., Liu, J., Wang, D., and Kang, Y.
\newblock Independent skill transfer for deep reinforcement learning.
\newblock In \emph{Proceedings of the Twenty-Ninth International Conference on
  International Joint Conferences on Artificial Intelligence (IJCNN)}, pp.\
  2901--2907, 2021{\natexlab{b}}.

\bibitem[Vashisht et~al.(2020)Vashisht, Rampal, Liao, Lu, Shanbhag, Fallon, and
  Kara]{vashisht2020placement}
Vashisht, D., Rampal, H., Liao, H., Lu, Y., Shanbhag, D., Fallon, E., and Kara,
  L.~B.
\newblock Placement in integrated circuits using cyclic reinforcement learning
  and simulated annealing.
\newblock \emph{arXiv preprint arXiv:2011.07577}, 2020.

\bibitem[Veli{\v{c}}kovi{\'c} et~al.(2017)Veli{\v{c}}kovi{\'c}, Cucurull,
  Casanova, Romero, Lio, and Bengio]{velivckovic2017graph}
Veli{\v{c}}kovi{\'c}, P., Cucurull, G., Casanova, A., Romero, A., Lio, P., and
  Bengio, Y.
\newblock Graph attention networks.
\newblock \emph{arXiv preprint arXiv:1710.10903}, 2017.

\bibitem[Viswanathan et~al.(2007{\natexlab{a}})Viswanathan, Nam, Alpert,
  Villarrubia, Ren, and Chu]{viswanathan2007rql}
Viswanathan, N., Nam, G.-J., Alpert, C.~J., Villarrubia, P., Ren, H., and Chu,
  C.
\newblock Rql: Global placement via relaxed quadratic spreading and
  linearization.
\newblock In \emph{Proceedings of the 44th annual Design Automation Conference
  (DAC)}, pp.\  453--458, 2007{\natexlab{a}}.

\bibitem[Viswanathan et~al.(2007{\natexlab{b}})Viswanathan, Pan, and
  Chu]{viswanathan2007fastplace}
Viswanathan, N., Pan, M., and Chu, C.
\newblock Fastplace 3.0: A fast multilevel quadratic placement algorithm with
  placement congestion control.
\newblock In \emph{2007 Asia and South Pacific Design Automation Conference
  (ASP-DAC)}, pp.\  135--140. IEEE, 2007{\natexlab{b}}.

\bibitem[Wang et~al.(2009)Wang, Chang, and Cheng]{wang2009electronic}
Wang, L.-T., Chang, Y.-W., and Cheng, K.-T.~T.
\newblock \emph{Electronic design automation: synthesis, verification, and
  test}.
\newblock Morgan Kaufmann, 2009.

\bibitem[Wang et~al.(2022{\natexlab{a}})Wang, Li, Wang, Kuang, Yuan, Zeng,
  Zhang, and Wu]{wang2022learning}
Wang, Z., Li, X., Wang, J., Kuang, Y., Yuan, M., Zeng, J., Zhang, Y., and Wu,
  F.
\newblock Learning cut selection for mixed-integer linear programming via
  hierarchical sequence model.
\newblock In \emph{The Eleventh International Conference on Learning
  Representations (ICLR)}, 2022{\natexlab{a}}.

\bibitem[Wang et~al.(2022{\natexlab{b}})Wang, Wang, Zhou, Li, and
  Li]{wang2022sample}
Wang, Z., Wang, J., Zhou, Q., Li, B., and Li, H.
\newblock Sample-efficient reinforcement learning via conservative model-based
  actor-critic.
\newblock In \emph{Proceedings of the AAAI Conference on Artificial
  Intelligence (AAAI)}, volume~36, pp.\  8612--8620, 2022{\natexlab{b}}.

\bibitem[Wang et~al.(2023)Wang, Pan, Zhou, and Wang]{wang2023efficient}
Wang, Z., Pan, T., Zhou, Q., and Wang, J.
\newblock Efficient exploration in resource-restricted reinforcement learning.
\newblock In \emph{Proceedings of the AAAI Conference on Artificial
  Intelligence (AAAI)}, volume~37, pp.\  10279--10287, 2023.

\bibitem[Xie et~al.(2021)Xie, Jiang, Wang, Xiong, and Bai]{xie2021policy}
Xie, T., Jiang, N., Wang, H., Xiong, C., and Bai, Y.
\newblock Policy finetuning: Bridging sample-efficient offline and online
  reinforcement learning.
\newblock \emph{Advances in neural information processing systems (NeurIPS)},
  34:\penalty0 27395--27407, 2021.

\bibitem[Yang et~al.(2000)Yang, Sarrafzadeh, et~al.]{yang2000dragon2000}
Yang, X., Sarrafzadeh, M., et~al.
\newblock Dragon2000: Standard-cell placement tool for large industry circuits.
\newblock In \emph{IEEE/ACM International Conference on Computer Aided Design.
  (ICCAD)}, pp.\  260--263. IEEE, 2000.

\bibitem[Zaruba \& Benini(2019)Zaruba and Benini]{zaruba2019cost}
Zaruba, F. and Benini, L.
\newblock The cost of application-class processing: Energy and performance
  analysis of a linux-ready 1.7-ghz 64-bit risc-v core in 22-nm fdsoi
  technology.
\newblock \emph{IEEE Transactions on Very Large Scale Integration (VLSI)
  Systems (TVLSI)}, 27\penalty0 (11):\penalty0 2629--2640, 2019.

\bibitem[Zheng et~al.(2022)Zheng, Zhang, and Grover]{zheng2022online}
Zheng, Q., Zhang, A., and Grover, A.
\newblock Online decision transformer.
\newblock \emph{arXiv preprint arXiv:2202.05607}, 2022.

\bibitem[Zhuang et~al.(2023)Zhuang, Lei, Liu, Wang, and
  Guo]{zhuang2023behavior}
Zhuang, Z., Lei, K., Liu, J., Wang, D., and Guo, Y.
\newblock Behavior proximal policy optimization.
\newblock \emph{arXiv preprint arXiv:2302.11312}, 2023.

\end{thebibliography}
\bibliographystyle{icml2023}

\newpage
\appendix
\onecolumn
\section{Appendix}

\subsection{Statistics of Benchmark Circuits}

In Table \ref{statistics} and \ref{statistics-industrial}, we provide the statistics of 26 circuits from benchmarks \textit{ISPD05}, \textit{ICCAD04}, and \textit{Ariane RISC-V CPU} and 6 {realistic} industrial circuits used in our experiments, where Macro (placed $1$st) means (the number of) macros placed by decision transformer.
For the three circuits \textit{adaptec1}, \textit{adaptec2}, and \textit{bigblue1}, because they have fixed macros that have been pre-placed around the chip as IOs, we maintain the positions of them but still compute HPWL with respect to these fixed macros to make a fair comparison. For circuits \textit{bigblue2} and \textit{bigblue4} that contain more than 8k macros, we select part of the macros to place based on the importance ordering. For circuits in benchmark \textit{ICCAD04}, the circuit description files do not distinguish between macros and standard cells. Thus we select 256 modules as macros and the remaining modules as standard cells.
In the implementation, we use the same circuit settings in all experiments.

\begin{table}[!ht]
 \small
	\caption{Statistics of public benchmark circuits.}
	\label{statistics}
	\centering
	\begin{tabular}{ccccccccc}
		\toprule
		Circuit & Macros & Macros (placed 1st)  & Hard Macros & Standard Cells & Nets & Pins & Ports & Area Util(\%)  \\
		\midrule
        adaptec1 & 543 & 63 & 63 & 210904 & 221142 & 944063 & 0 & 55.62  \\ 
        adaptec2 & 566 & 159 & 159 & 254457 & 266009 & 1069482 & 0 & 74.46  \\ 
        adaptec3 & 723 & 723 & 201 & 450927 & 466758 & 1875039 & 0 & 61.51  \\ 
        adaptec4 & 1329 & 1329 & 92 & 494716 & 515951 & 1912420 & 0 & 48.62  \\ 
        bigblue1 & 560 & 32 & 32 & 277604 & 284479 & 1144691 & 0 & 31.58  \\ 
        bigblue2 & 23084 & 256 & 52 & 534782 & 577235 & 2122282 & 0 & 32.43  \\ 
        bigblue3 & 1293 & 1293 & 138 & 1095519 & 1123170 & 3833218 & 0 & 66.81  \\ 
        bigblue4 & 8170 & 1024 & 52 & 2169183 & 2229886 & 8900078 & 0 & 35.68  \\ 
        ariane & 932 & 932 & 134 & 0 & 12404 & 22802 & 1231 & 78.39  \\ 
        ibm01 & 256 & 256 & 52 & 12506 & 14111 & 50566 & 246 & 61.94  \\ 
        ibm02 & 256 & 256 & 52 & 19321 & 19584 & 81199 & 259 & 64.63  \\ 
        ibm03 & 256 & 256 & 52 & 22846 & 27401 & 93573 & 283 & 57.97  \\ 
        ibm04 & 256 & 256 & 52 & 26899 & 31970 & 105859 & 287 & 54.88  \\ 
        ibm06 & 256 & 256 & 52 & 32320 & 34826 & 128182 & 166 & 54.77  \\ 
        ibm07 & 256 & 256 & 52 & 45419 & 48117 & 175639 & 287 & 46.03  \\ 
        ibm08 & 256 & 256 & 52 & 51000 & 50513 & 204890 & 286 & 47.13  \\ 
        ibm09 & 256 & 256 & 52 & 53142 & 60902 & 222088 & 285 & 44.52  \\
        ibm10 & 256 & 256 & 52 & 68643 & 75196 & 297567 & 744 & 61.40  \\ 
        ibm11 & 256 & 256 & 52 & 70185 & 81454 & 280786 & 406 & 41.40  \\ 
        ibm12 & 256 & 256 & 52 & 70425 & 77240 & 317760 & 637 & 53.85  \\ 
        ibm13 & 256 & 256 & 52 & 83775 & 99666 & 357075 & 490 & 39.43  \\ 
        ibm14 & 256 & 256 & 52 & 146991 & 152772 & 546816 & 517 & 22.49  \\ 
        ibm15 & 256 & 256 & 52 & 161177 & 186608 & 715823 & 383 & 28.89  \\ 
        ibm16 & 256 & 256 & 52 & 183026 & 190048 & 778823 & 504 & 39.46  \\ 
        ibm17 & 256 & 256 & 52 & 184735 & 189581 & 860036 & 743 & 19.11  \\ 
        ibm18 & 256 & 256 & 52 & 210328 & 201920 & 819697 & 272 & 11.09 \\ 
		\bottomrule
	\end{tabular}
	
\end{table}

\begin{table}[!ht]
 \small
	\caption{Statistics of industrial benchmark circuits.}
	\label{statistics-industrial}
	\centering
	\begin{tabular}{ccccccc}
		\toprule
		Circuit & Macros & Macros (placed 1st) & Standard Cells & Nets & Pins & Ports  \\
		\midrule
        C1 & 45 & 45 & 652519 & 817469 & 2792845 & 9009  \\ 
        C2 & 159 & 159 & 3487686 & 4417159 & 9532116 & 6676  \\ 
        C3 & 134 & 134 & 2268372 & 2776101 & 10004114 & 1416  \\ 
        C4 & 70 & 70 & 1058352 & 1340788 & 4256481 & 27120  \\ 
        C5 & 144 & 144 & 2766094 & 3143780 & 11519226 & 11174 \\ 
        C6 & 165 & 165 & 1777410 & 2443917 & 7268601 & 14882\\
		\bottomrule
	\end{tabular}

\end{table}

\subsection{Additional Results} 
\begin{figure*}[!htbp]
\centering
\includegraphics[width=\columnwidth]{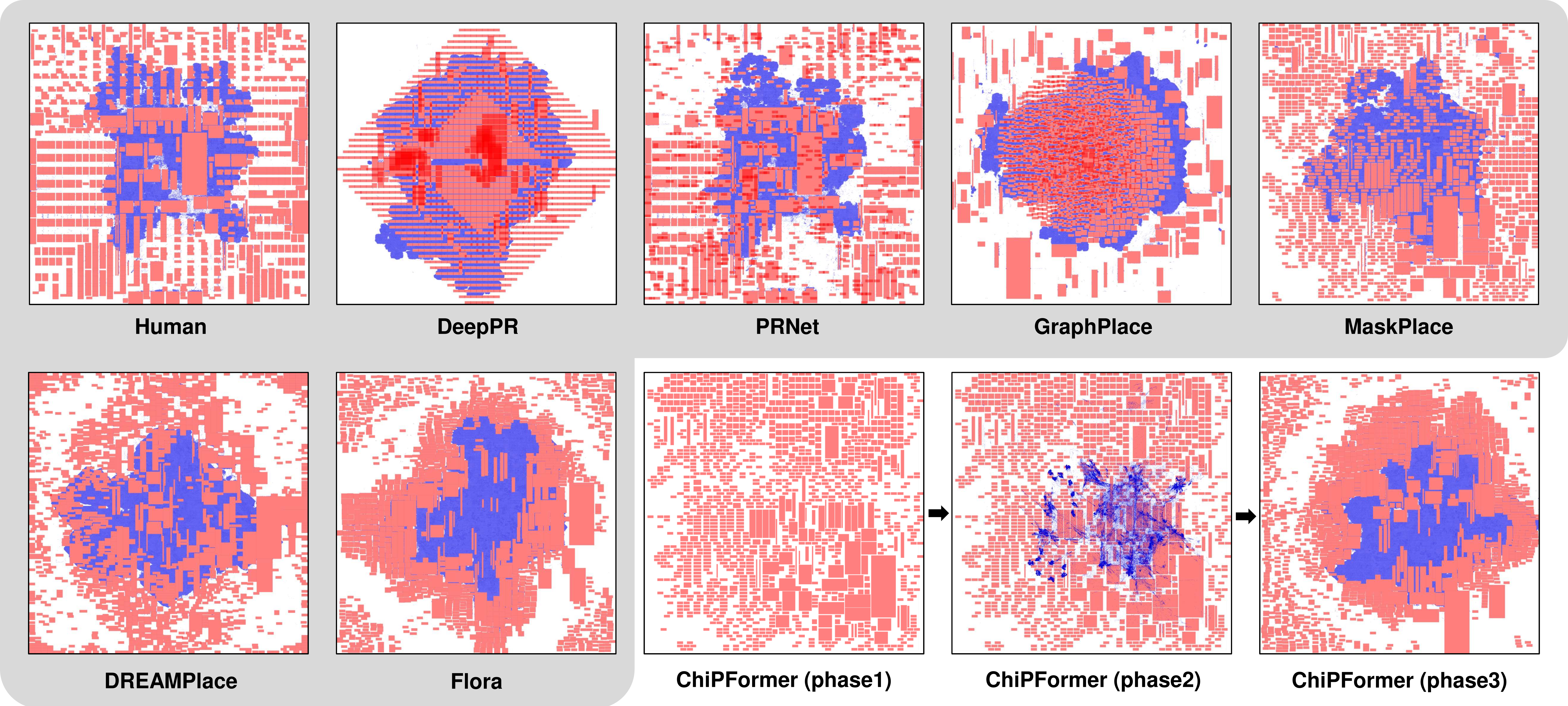}
\caption{\textbf{Visualization of mixed-size placement for circuit \textit{adaptec4}.} Results include Human (HPWL = 17.44$\times$$10^7$, Overlap = 0\%), DeepPR (HPWL = 23.40$\times$$10^7$, Overlap = 32.11\%), PRNet (HPWL = 23.64$\times$$10^7$, Overlap = 13.36\%), GraphPlace (HPWL = 25.58$\times$$10^7$, Overlap = 7.43\%), MaskPlace (HPWL = 22.97$\times$$10^7$, Overlap = 0\%), DREAMPlace (HPWL = 14.41$\times$$10^7$, Overlap = 0\%), Flora (HPWL = 14.30$\times$$10^7$, Overlap = 0\%), and ChiPFormer (HPWL = \underline{12.97$\times$$10^7$}, Overlap = \underline{0\%}).}
\label{visualization_app}
\end{figure*}

\begin{figure}
	\centering
	\includegraphics[width=0.3\columnwidth]{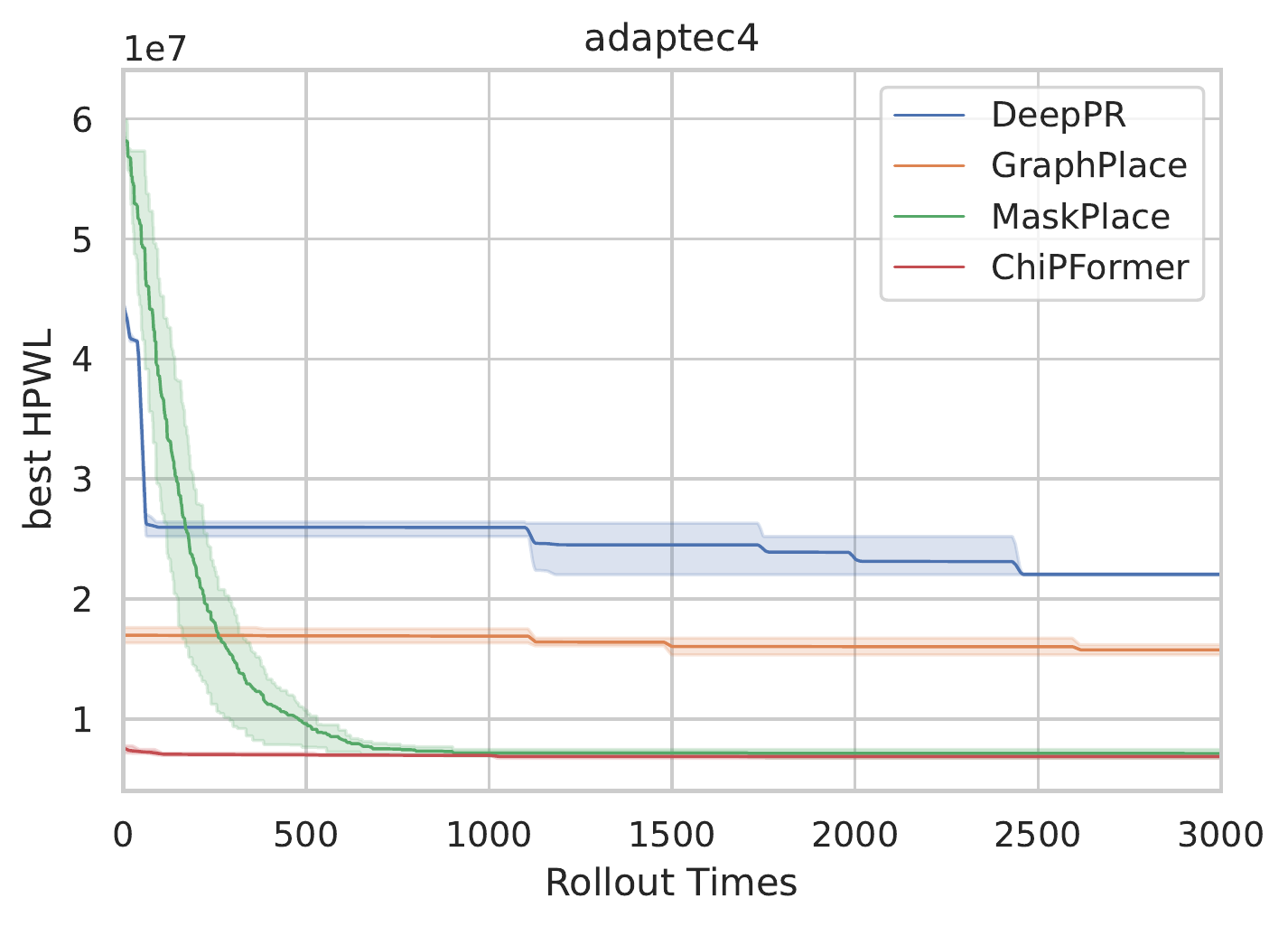}
	\includegraphics[width=0.3\columnwidth]{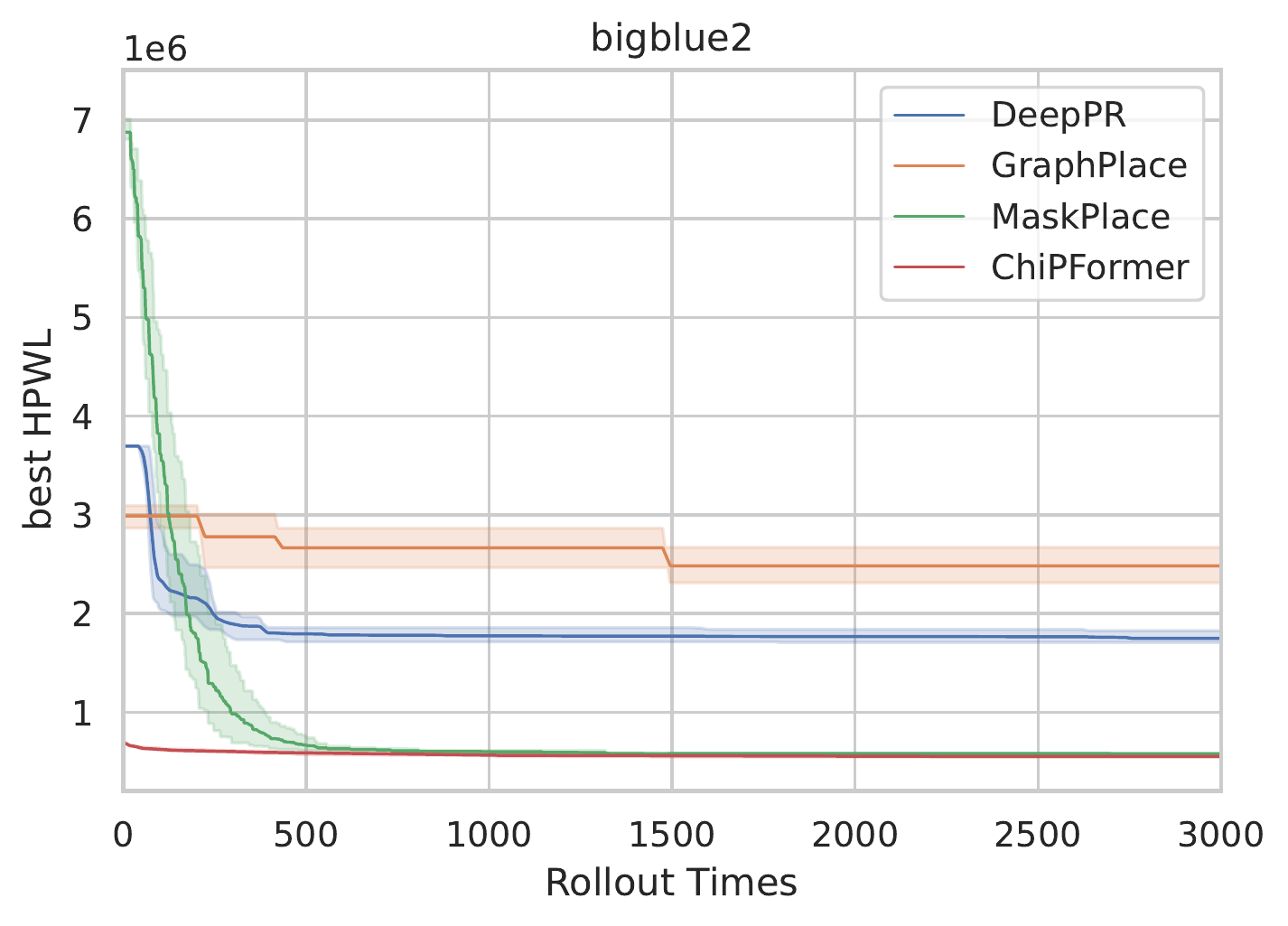}
	\caption{\textbf{HPWL versus rollout times (number of trajectories).} ChiPformer can reduce the rollout times by more than 90\% while achieving the same placement quality. }
	\label{traj-vs-hpwl}
\end{figure}

\begin{figure*}[!htbp]
\centering
\includegraphics[width=\columnwidth]{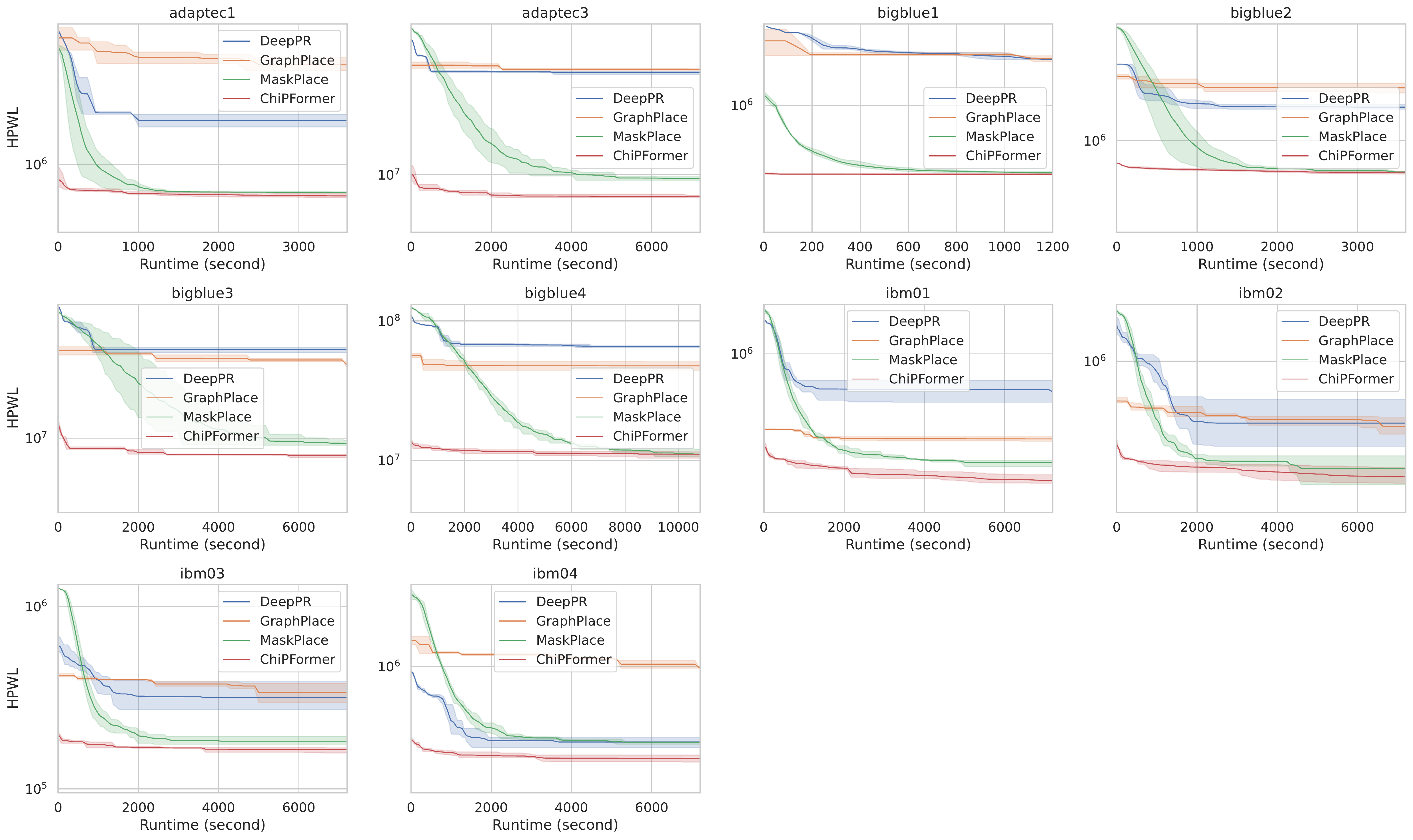}
\caption{\textbf{HPWL curves over runtime. (cont.)} All methods are evaluated on two RTX3090 GPUs.}
\label{hpwl-versus-time}
\end{figure*}

In Fig.\ref{visualization_app}, we provide more visualization results for mixed-size placement. 
In Fig.\ref{traj-vs-hpwl}, we provide additional results (on circuit \textit{adaptec4} and \textit{bigblue2}) for the training sample efficiency comparison. 
In Fig.\ref{hpwl-versus-time}, we present additional results (over benchmarks \textit{ISPD05} and \textit{ICCAD04}) for the training time efficiency comparison. 
In Table \ref{overlap}, we provide the comparison results of the overlap ratio for macro placement. 
In Table \ref{full-placement-iccad}, we give additional results for the mixed-size placement in \textit{ICCAD04} benchmark. 
In Table \ref{reconstruct}, we give out the reconstruct quality of testing set and the related code is from original repository \footnote{\href{github.com/DaehanKim/vgae\_pytorch}{https://github.com/DaehanKim/vgae\_pytorch}}.

\begin{table}[!htb]
	\caption{
		\textbf{Comparisons of the overlap ratio (\%) for macro placement}, which are evaluated by the ratio of the cumulative overlapping area to the total area in the chip. The overlap ratio should be 0 (or as close to 0 as possible) to meet the requirements of chip manufacturing. We can observe that ChipFormer can maintain non-overlapping placements in most circuits.
	}
	\label{overlap}
	\begin{center}
		\begin{adjustbox}{max width=\columnwidth}
			\begin{tabular}{cccccc}
				\toprule circuit & GraphPlace & DeepPR & PRNet & MaskPlace & ChiPFormer \\
				\midrule
				adaptec1  & 1.89 & 46.26 & 12.60 & {0.00} & {0.00}\\
				adaptec2   & 1.54 & 18.39 & 9.94 & {0.00} & {0.00}\\
				adaptec3   & 1.24& 24.35 & 10.71&  {0.00} & {0.00}\\
				adaptec4   & 7.59 & 32.11 & 13.36 & {0.00} & {0.00}\\
				bigblue1   & 1.98 & 2.08 & 2.04 & {0.00} & {0.00}\\
				bigblue2  & 0.77 & 29.42 & 20.06 & {0.00} & {0.00}\\
				bigblue3  & 0.96 & 74.03 & 5.11 & {0.00} & {0.00}\\
				bigblue4  & 5.54 & 8.35 & 6.46 & {0.00} & {0.00}\\
				ariane & 5.13 & 38.91 & 9.49 & {3.33} & {3.27}\\
				\bottomrule
			\end{tabular}
		\end{adjustbox}
	\end{center}
\end{table}

\begin{table}[!htb]
	\caption{\textbf{Comparisons of HPWL ($\times10^7$) for \textbf{mixed-size} placement in \textit{ICCAD04} benchmark.} HPWL is the smaller the better. SA (simulated annealing) method is implemented as in \citet{mirhoseini2021graph}. ChiPFormer(workflow) can achieve the best placement quality.}
	\label{full-placement-iccad}
	\begin{center}
		\begin{adjustbox}{max width=\columnwidth}
			\begin{tabular}{cccccc}
				\toprule circuit & SA & RePlAce & GraphPlace & MaskPlace & ChiPFormer(workflow) \\
				\midrule
				ibm01  & 25.85 & \underline{22.82} & 31.71 & 24.18 & \textbf{16.70}  \textcolor{magenta}{\scriptsize{(-26.82\%)}}\\
				ibm02  & 54.87 & 47.59 & 55.11 & \underline{47.45} & \textbf{37.87} \textcolor{magenta}{\scriptsize{(-20.19\%)}}\\
				ibm03  & 80.68 & \underline{64.36} & 80.00 & 71.37 & \textbf{57.63} \textcolor{magenta}{\scriptsize{(-10.46\%)}}\\
				ibm04  & 83.32 & \underline{72.61} & 86.86 & 78.76 & \textbf{65.27} \textcolor{magenta}{\scriptsize{(-10.11\%)}}\\
				ibm06 & 69.09 & 58.07 & 63.48 & \underline{55.70} & \textbf{52.57} \textcolor{magenta}{\scriptsize{(-5.62\%)}}\\
				ibm07 & 111.03 & 98.57 & 117.70 & \underline{95.27} & \textbf{86.20} \textcolor{magenta}{\scriptsize{(-9.52\%)}}\\
				ibm08 & 131.07 & \underline{114.67} & 134.77 & 120.64 & \textbf{102.26} \textcolor{magenta}{\scriptsize{(-10.82\%)}}\\
				ibm09 & 135.45 & \underline{120.01} & 148.74 & 122.91 & \textbf{105.61} \textcolor{magenta}{\scriptsize{(-12.00\%)}}\\
				ibm10 & 423.14 & \underline{274.29} & 440.78 & 367.55 & \textbf{230.39} \textcolor{magenta}{\scriptsize{(-16.00\%)}}\\
				ibm11 & 210.12 &  \underline{169.98} & 218.73 & 202.23 & \textbf{160.60} \textcolor{magenta}{\scriptsize{(-5.52\%)}} \\
				ibm12 & 410.05 & \underline{306.33} & 438.57 & 397.25& \textbf{273.14} \textcolor{magenta}{\scriptsize{(-10.83\%)}} \\
				ibm13 & 259.89 & \underline{220.14} & 278.92 & 246.49 & \textbf{197.20} \textcolor{magenta}{\scriptsize{(-10.42\%)}}\\
				ibm14 & 405.80 & 341.80 & 455.32 & \underline{302.67} & \textbf{301.28} \textcolor{magenta}{\scriptsize{(-0.46\%)}}\\
				ibm15 & 510.06 & \underline{451.36} & 520.06 & 457.86 & \textbf{429.71} \textcolor{magenta}{\scriptsize{(-4.80\%)}}\\
				ibm16 & 614.54 & \underline{516.05} & 642.08 & 584.67 & \textbf{463.32} \textcolor{magenta}{\scriptsize{(-10.22\%)}}\\
				ibm17 & 720.40 & \underline{635.93} & 814.37 & 643.75 & \textbf{569.13} \textcolor{magenta}{\scriptsize{(-10.50\%)}} \\
				ibm18 & 442.00 & 399.43 & 450.67 & \underline{398.83} & \textbf{370.36} \textcolor{magenta}{\scriptsize{(-7.14\%)}} \\
				\bottomrule
			\end{tabular}
		\end{adjustbox}
	\end{center}
\end{table}

\begin{table}[!htb]
	\caption{\textbf{Reconstruct quality of VGAE.} AUC, AP, and ACC mean area under the receiver operating characteristic curve, average precision from prediction scores, and accuracy, respectively.}
	\label{reconstruct}
	\begin{center}
			\begin{tabular}{cccc}
   \toprule
                    Test Metrics & AUC & AP & ACC \\
                    \midrule
                    Value & 0.9600 & 0.9557 & 0.8969 \\
				\bottomrule
			\end{tabular}
	\end{center}
\end{table}

\begin{figure}[!htb]
	\vskip 0.2in
	\begin{center}
		\centerline{\includegraphics[width=0.95\columnwidth]{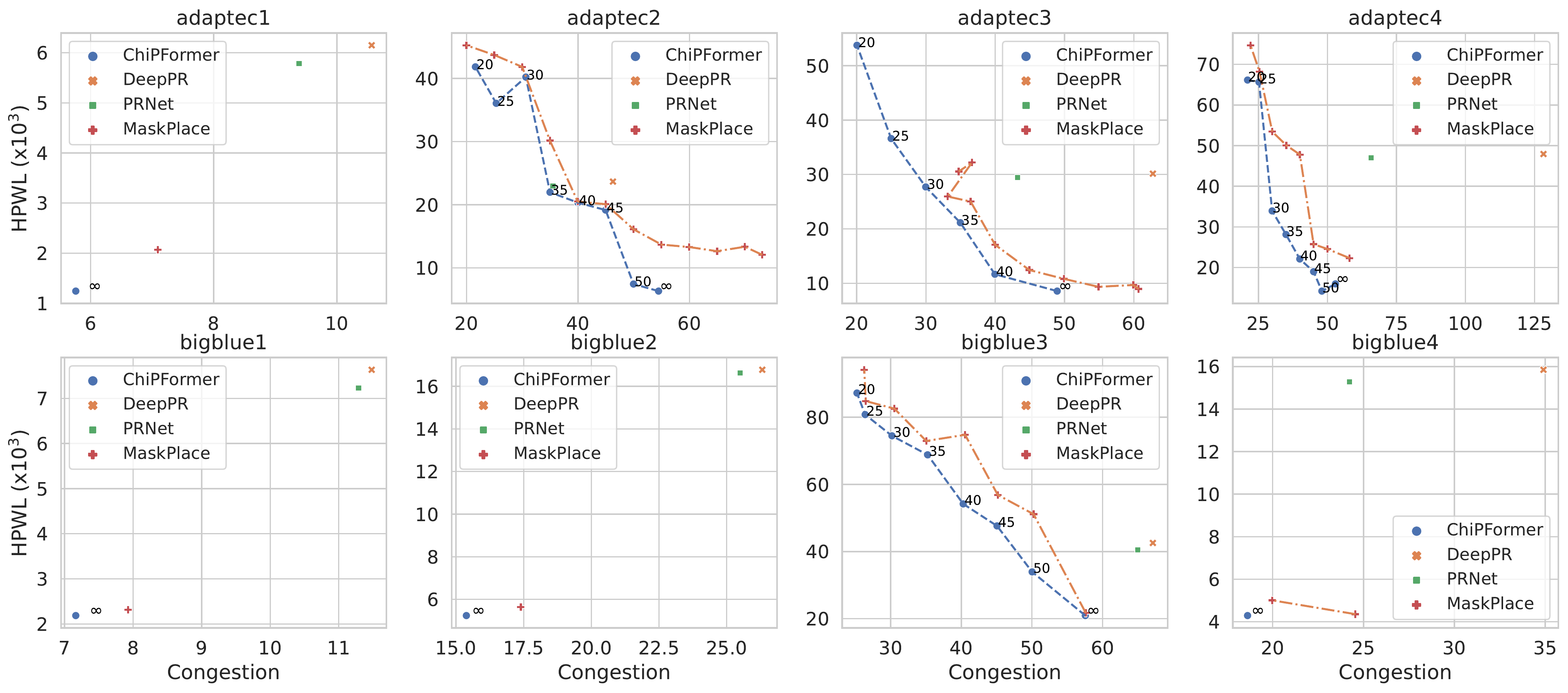}}
		\caption{\textbf{Comparisons over HPWL and congestion.} Both HPWL and congestion are smaller the better. Numbers beside points are the expected congestion thresholds $C$ (as the input parameter). We run ChiPFormer finetuning with 300 rollouts, then set the congestion threshold to $+\infty$ and gradually decrease it.
		}
		\label{congestion}
	\end{center}
	\vskip -0.2in
\end{figure}

\textbf{Congestion Constraint.} 
Considering that some scenarios require the congestion metric, we test ChiPFormer on both HPWL and congestion metrics. We run ChiPFormer on 8 circuits in \textit{ISPD05}. To make a fair comparison, we run experiments on the processed circuits from PRNet \cite{chengpolicy}. 
We show the results in Fig.\ref{congestion}. At testing, we first set the congestion threshold to $+\infty$ and find how far congestion can go when considering only HPWL. Then, we gradually reduce this threshold to find all points on the Pareto front. We can find that our ChiPFormer can achieve better results in both HPWL and congestion metrics when compared to MaskPlace, DeepPR and PRNet.

\subsection{Example on the Circuit Token}
\label{didactic-example}
\begin{figure}[!htb]
\begin{center}
\centerline{\includegraphics[width=0.7\columnwidth]{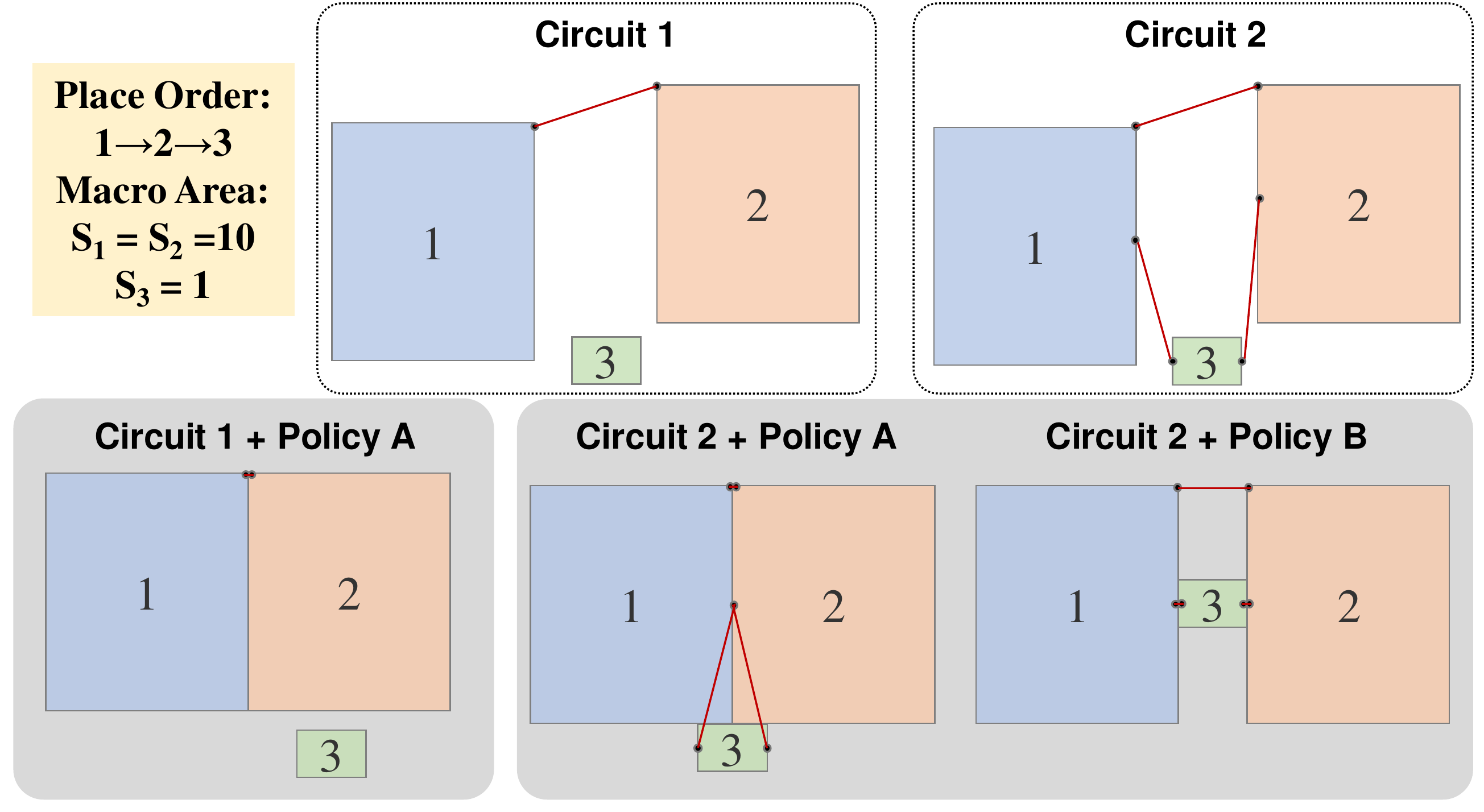}}
\caption{\textbf{An example for two circuits that share the same state and action tokens ($\bs_1, \ba_1, \bs_2, \ba_2$).} Consider the two circuits (circuit 1 and circuit 2). The only difference between them is that the 3rd macro of the circuit is connected to both the 1st and 2nd macros in circuit 2. According to the definition of state token, the state tokens $\bs_1, \bs_2$ are the same because the position, wire, and view masks only involve the placed macros and the current macro to be placed. That means state $s_1$ contains the 1st macro and state $\bs_2$ contains the 1st and 2nd macros, which are the same in two circuits.
Consider an optimal policy A that minimizes the HPWL in circuit 1 (subplot ``Circuit 1 + Policy A''). However, deploying policy A over circuit 2 will lead to the sub-optimal solution (subplot ``Circuit 2 + Policy A'').
In contrast, the optimal policy for circuit 2 should be Policy B, which renders different behaviors at state $\bs_2$ (subplot ``Circuit 2 + Policy B''). Thus, in the multi-task chip placement setting, the circuit token helps us identify optimal placement behaviors when circuits contain the same macros while differing in their topology structures.}
\label{circuit_emb}
\end{center}
\end{figure}

The agent learns policy only from state and action tokens when ablating the circuit token (\ie ignoring the topology structure of the circuit). 
Suppose two circuits contain the same modules but are connected differently. In that case, the agent can not distinguish them because they share the same state and action tokens and thus generate the same actions in the two circuits.
However, that will lead to a sub-optimal solution on one circuit.
To illustrate, we give an example in Fig.\ref{circuit_emb}.

\subsection{State Token Generation}
\label{state-token}
\begin{figure}[!htb]
\begin{center}
\centerline{\includegraphics[width=\columnwidth]{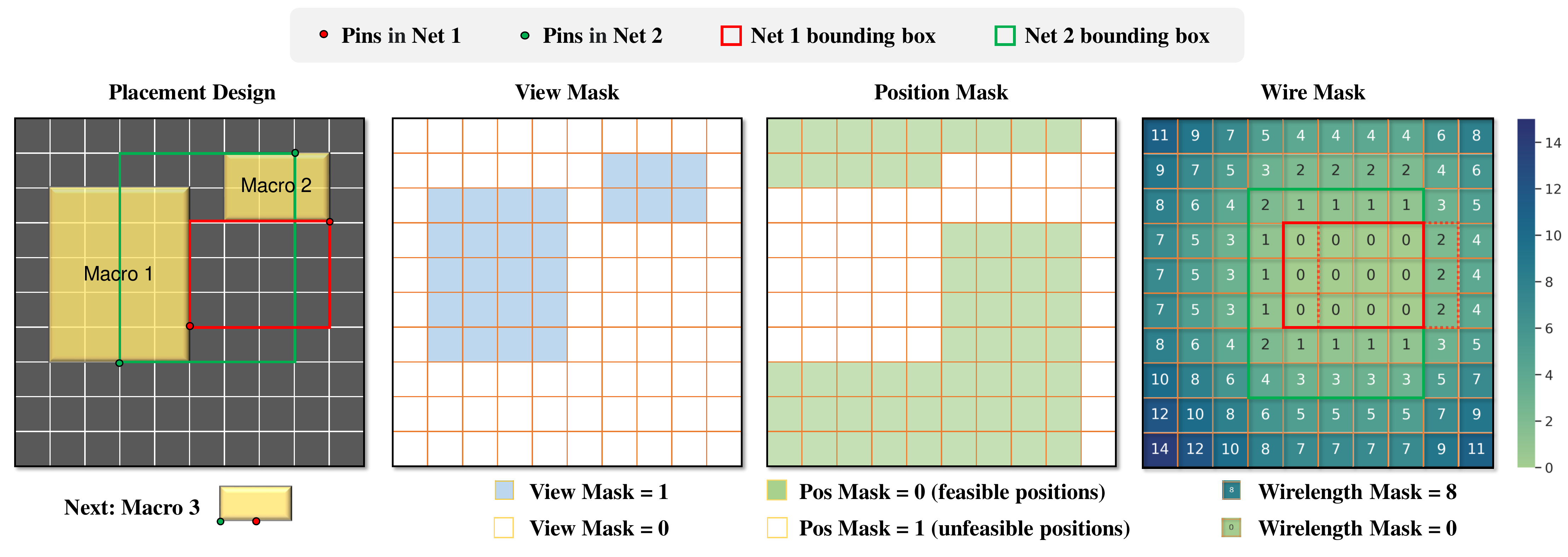}}
\caption{\textbf{An example for state tokens (placement masks) generation.} The view mask, position mask, and wire mask are generated based on the current placement design and the next macro to place. View mask $V\in \{0,1\}^{N\times N}$ is to mark the occupied cells by macros. Position mask $P\in \{0,1\}^{N\times N}$ is to label all feasible action positions that do not cause overlap. Wire mask $W\in [0,1]^{N\times N}$is to label the HPWL increase when the next macro is placed in the corresponding position. For ease of understanding, we marked the data before normalization for the wire mask, which is the sum of the shortest Manhattan distances for each cell to the inside of bounding boxes.}
\label{mask}
\end{center}
\end{figure}

The state tokens in ChiPFormer consist of the view mask, position mask, and wire mask as in MaskPlace \cite{lai2022maskplace}, which are representations of the placement state from different perspectives. Each mask occupies one input channel. All masks can be seen as images with resolution $N \times N$. A state token generation example can be seen in Fig.\ref{mask}. In this case, there are already two placed macros, and now the agent will take action to place Macro 3. There are two nets, and the corresponding pins and bounding boxes are marked in \textcolor{red}{red} and \textcolor{green}{green}, respectively. 
(1) View mask marks all occupied grid cells by macros with 1. 
(2) Position mask labels all feasible cells that will not generate an overlap if the next macro is placed at the position (supposed that the placement position corresponds to the lower-left corner of the macro). 
(3) Wire mask marks the HPWL increase when the next macro is placed in the corresponding cells. According to the definition of HPWL, the increase in HPWL is the increase in the size of the net bounding box. Thus, if the place position is inside the net bounding box, the HPWL increase equals 0. Otherwise, it equals the distance from the cells to the bounding box. It is easy to find that the values in the wire mask are the sum of the distances from the cells to all net bounding boxes as Fig.\ref{mask}. Considering that some pins are not located in the lower-left corner of the macro (an offset), the corresponding net bounding box needs to be moved by an equivalent offset (\ie the bounding box is moved in the opposite direction of the offset vector).

\subsection{Model Architecture and Hyper-parameters}
\label{architecture-and-hyper}
The detailed model architecture can be seen in Table \ref{model-arch-tab}. Hyper-parameters of the ChiPFormer can be found in Table \ref{configuration-tab}. When pretraining, we keep the sequence length $T$=256. If a circuit contains more than 256 macros, the trajectory is truncated to the beginning of 256 macros according to the placement order. On the contrary, if there are less than 256 macros, we pad the trajectory with zeros to the fixed length. When testing, for those circuits with more than $T=256$ macros, ChiPFormer keeps a sliding window to record the latest 256 state tokens and 256 action tokens to generate the next action, following the way of the decision transformer.

\begin{table}[!htb]
	\caption{Model Architecture}
	\label{model-arch-tab}
    \centering
    \begin{tabular}{c|c|c|c}
    \toprule
         ~ & layer name (index) & kernel size & output size  \\ \midrule
        \multirow{3}{*}{\shortstack[c]{circuit token extraction\\(VGAE)}} & GCN\_1 & - & (num of macros, 32)  \\ 
        ~ & GCN\_2 & - & (num of macros, 32)  \\ 
        ~ & Pooling & - & (32, )  \\ \midrule
        circuit embedding  & FC\_1 & - & (1024,)  \\ 
        ~ & FC\_2 & - & (1024,)  \\ 
        ~ & FC\_3 & - & (768, )  \\ \midrule
        state embedding & CNN\_1 & 8×8, 16 & (16, 40, 40)  \\ 
        ~ & CNN\_2 & 4×4, 32 & (32, 20, 20)  \\ 
        ~ & CNN\_3 & 3×3, 16 & (16, 10, 10)  \\ 
        ~ & FC\_1 & - & (784,)  \\ \midrule
        action embedding & Embedding & - & (7056,)  \\  \midrule
        action head & reshape to 2D & - & (84, 84)  \\ 
        ~ & CNN\_1 & 1×1, 8 & (8, 84, 84)  \\ 
        ~ & CNN\_2 & 1×1, 8 & (8, 84, 84)  \\ 
        ~ & CNN\_3 & 1×1, 1 & (84, 84)  \\ \midrule
        action merge & CNN\_1 & 1×1, 1 & (84, 84) \\ 
        \bottomrule
    \end{tabular}
\end{table}

\begin{table}[!htb]
	\caption{Hyper-parameters used in our experiments}
	\label{configuration-tab}
    \centering
    \small
    \begin{tabular}{c|cc|cc}
    \toprule
        Stage & Configuration & Value &Configuration & Value\\ \midrule
        \multirow{3}{*}{circuit token extraction} & dim of hidden layer & 32 & dim of latent variable & 32  \\ 
        ~ &training iterations & 800 & learning rate & 1e-2\\ 
        ~ &num of node features & 4 & ~ & ~\\ \midrule
         \multirow{3}{*}{pretraining} & num of layers \#L & 6 & hidden size \#H & 128  \\ 
        ~ &num of att heads \#A & 8 & num of tokens & 1+2$\times$256\\
        ~ & batch size & 32 & learning rate & 6e-4\\ \midrule
         \multirow{3}{*}{finetuning} & temperature for sampling $\alpha$ & 1e-2 & replay buffer size  & 64  \\ 
         ~ & entropy weight $\lambda$ & 0.5 & expected entropy $\beta$& 0.5\\
        ~ &batch size & 32 & learning rate & 1e-4\\ \midrule
        \multirow{2}{*}{standard cell placement} & density weight in (phase 2)& 1e-4 & density weight in (phase 3)& 1e-3 \\
        ~ & number of iterations (phase 2) & 300\\ \midrule
         computing hardware & CPU & AMD Ryzen 9 5950X&  GPU & 2$\times$ RTX 3090\\
        \bottomrule
    \end{tabular}
\end{table}

\subsection{Pseudo-code for Circuit Token Generation}
The pseudo-code version for generating circuit token $\text{HI}(c_{test})$ of circuit $c_{\text{test}}$ can be seen in Algo. \ref{alg-circuit-token}.

\section{Related Work}
\textbf{Optimization-based Method for Placement.}
As a combinatorial optimization problem, researchers have tried different optimization-based methods to solve the chip placement task, which can be classified into partitioning-based, simulated annealing-based, and analytic-based methods. 

Partitioning-based methods \cite{roy2006min, khatkhate2004recursive} are based on the divide-and-conquer idea. They first divide the circuits into sub-circuits and assign each part to a sub-region on the chip. Then, these sub-circuits can be continuously divided and assigned to divided sub-regions recursively. These methods are efficient but can hardly get high placement quality, especially for large-scale circuits.

Simulated annealing-based methods \cite{yang2000dragon2000, vashisht2020placement} start from a randomly generated placement solution and then try to move to a neighbor solution with a slight change. If the neighbor solution is better than the current solution, it uses the neighbor solution to replace it. If the neighbor solution is worse than the current solution, there is also a probability of decreasing with the iterative process to accept the neighbor solution. Such a solution can achieve improved placement quality but suffers from high computational inefficiency.

Analytical-based methods express the optimization problem as an analytical function of the coordinates of modules. Based on the analytical functions, these methods can be divided into quadratic methods \cite{viswanathan2007rql, viswanathan2007fastplace, kim2012maple, kim2012complx, brenner2015bonnplace, lin2013polar, spindler2008kraftwerk2} and non-quadratic methods \cite{chen2008ntuplace3, lu2014eplace, cheng2018replace, lin2020dreamplace, chan2006mpl6, kahng2005aplace,  gu2020dreamplace}. Quadratic methods use quadratic functions as objective functions and solve placement by Quadratic Programming (QP) \cite{nocedal2006quadratic}. These methods are more applicable to problems where the search space is discrete, so they are still used in FPGA placement. On the contrary, in the non-quadratic methods, the objective functions are not quadratic and contain some non-linear functions, such as the logarithmic function. The advantage is that these functions can describe the placement target more precisely \cite{cheng2018replace, lu2014eplace}. However, the corresponding optimization methods, such as stochastic gradient descent, will lead to a locally optimal solution. 

\textbf{Learning-based Method for Placement.} 
Deep learning-based methods \cite{wang2022sample,wang2023efficient} have been used to solve tasks in various fields, such as mathematics \cite{davies2021advancing}, biomedical science \cite{jumper2021highly}, combinatorial optimization \cite{wang2022learning} and information security \cite{lai2020opensmax}.
With the development of deep learning techniques, researchers have proposed learning-based placement methods, which can be divided into reinforcement learning-based, supervised learning-based, and unsupervised learning-based methods.

All existing reinforcement learning-based methods belong to online methods. They formulate the placement as a sequential Markov decision process and decide the position of one module at each step. GraphPlace \cite{mirhoseini2021graph} uses the policy gradient framework \cite{sutton1999policy} to encode the current placement state by GCN-based model and generates the probability matrix of placement positions by de-convolution layers. After placing all hard macros, it uses the force-directed method (a quadratic method) \cite{spindler2008kraftwerk2} to place the remaining soft macros and standard cells. After that, \citet{jiang2021delving} verified that using the non-quadratic method DREAMPlace rather than the force-directed method could achieve better results. DeepPR \cite{cheng2021joint} and PRNet \cite{chengpolicy} combine the CNN and GCN methods to encode the placement state and train the PPO model \cite{schulman2017proximal} to learn the policy. However, they do not consider the actual sizes of macros, which means the reward function for HPWL is not accurate, and the overlaps cannot be avoided when placing macros. MaskPlace \cite{lai2022maskplace} introduces a series of visual representations of states in placement from the perspective of view, position, and wirelength. All of them are 2D images, where the view image is a high-level representation to describe the placement status from the human view. The position and wirelength images are low-level representations of overlapping positions and wirelength increases. 
However, it does not consider the topology information of circuits and previous states when making decisions by the CNN-based model, which tends to lead to a sub-optimal solution.
\citet{kimcollaborative} also proposed an offline reinforcement learning method for the recap placement task, but it is for high-frequency analog circuits, the scale of which is much smaller than the placement of large-scale digital circuits.

Supervised learning-based methods such as Flora \cite{liu2022floorplanning} and GraphPlanner \cite{liu2022graphplanner} considered that transforming the placement task into a sequential decision process makes itself complicated. On the contrary, they train the GAT \cite{velivckovic2017graph} and Variation GCN \cite{kipf2016semi} models on a synthetic training dataset to predict the positions of modules as a kind of prediction task. However, the prediction results are hardly used directly because the non-overlapping constraint is not considered. Also, the synthetic data is still far from the real circuits.

Unsupervised learning-based methods are still in the early stages of research. \citet{lu2022placement} proposed to use the GNN-based model to cluster modules and use the clustering results as soft constraints for the following placement solvers (\ie the solver will make the positions of modules under the same cluster as close as possible). Due to a lack of labeling information for clustering, the estimated objective function based on metrics such as congestion, timing, and power is not guaranteed to be accurate.

\textbf{Offline Reinforcement Learning.} Currently,  the task of (offline) RL covers a wide range of problem settings: from the usual single-domain setting~\citep{levine2020offline} to cross-domain settings~\citep{liu2021unsupervised, liudara2022}, from offline training~\citep{fujimoto2019off, zhuang2023behavior} to online finetuning settings~\citep{xie2021policy, lee2022offline}, from unsupervised learning skills~\citep{chebotar2021actionable, tian2021unsupervised, tian2021independent, liu2022learn} to learning from human feedback~\citep{shin2021offline, kang2023beyond}. Consistently, these works consider simulated environments (or simulated D4RL datasets~\citep{fu2020d4rl}). In this paper, however, we focus on algorithms and implementations for offline RL in real chip placement tasks. We believe this can help inspire more deployments of RL algorithms to real-world tasks.

{\centering
\begin{minipage}{.6\linewidth}
\begin{algorithm}[H]
   \caption{Generate circuit token $\text{HI}(c_{test})$}
   \label{alg-circuit-token}
\begin{algorithmic}[1]
   \STATE {\bfseries Input:} Training data $\mathcal{M}_{\text{train}}^{[n]}$, including adjacency metrics $A^{[n]}$, node features $X^{[n]}$ of $n$ chip circuits, inference model $q_{\phi}(X,A)$, and generative model $p(Z)$. Testing data $\mathcal{M}_{\text{test}}$ for the unseen circuit $c_{\text{test}}$.
   \FOR{$\text{epoch}=1$ {\bfseries to} $E$}
   \STATE $Z^{[n]} =q_{\phi}(X^{[n]},A^{[n]})$
   \STATE $\hat{A}^{[n]} = p(Z^{[n]})$
   \STATE $\text{loss} = \text{BCEloss}(A^{[n]}, \hat{A}^{[n]})$
   \STATE $\phi = \phi - \partial \text{loss}/\partial \phi$
   \ENDFOR
   \STATE $Z =q_{\phi}(X^{c_{\text{test}}},A^{c_{\text{test}}})$
   \COMMENT{\textcolor{blue}{Generate latent variables for N macros.}}
   \STATE $\text{HI}(c_{\text{test}}) = \sum_{i=1}^N{z_i}/N$ 
   \COMMENT{\textcolor{blue}{Circuit token (chip representation) is the average pooling for all variables of macros}}
\end{algorithmic}
\end{algorithm}
\end{minipage}
\par
}

\end{document}